\documentclass[12pt,letterpaper]{article}
\usepackage{amssymb,amsmath}
\usepackage{mathtools}
\usepackage{amsthm}
\usepackage{bbm}
\usepackage{graphicx}
\usepackage{enumerate}
\usepackage{caption}
\usepackage{natbib}
\usepackage{url} 
\usepackage{bm}
\usepackage{color}
\usepackage{placeins} 
\usepackage{silence} 
\DeclareMathOperator*{\median}{\mbox{median}}
\DeclareMathOperator{\mad}{\mbox{mad}}

\DeclareMathOperator*{\argmax}{\mbox{argmax}}

\newcommand{\bv}{\boldsymbol v}
\newcommand{\bx}{\boldsymbol x}
\newcommand{\bt}{\boldsymbol t}
\newcommand{\btx}{\boldsymbol{\tilde{x}}}

\newcommand{\bV}{\boldsymbol V}

\newcommand{\bbeta}{\boldsymbol \beta}

\newcommand{\bhbeta}{\boldsymbol{\hat{\beta}}}

\newcommand{\bmu}{\boldsymbol \mu}
\newcommand{\bhmu}{\boldsymbol{\hat{\mu}}}
\newcommand{\bSigma}{\boldsymbol \Sigma}
\newcommand{\bhSigma}{\boldsymbol{\widehat{\Sigma}}}

\newcommand{\hf}{\hat{f}}
\newcommand{\hg}{\hat{g}}
\newcommand{\tg}{\tilde{g}}

\newcommand{\hp}{\hat{p}}
\newcommand{\tp}{\tilde{p}}
\newcommand{\hy}{\hat{y}}
\newcommand{\hbeta}{\hat{\beta}}
\newcommand{\hpi}{\hat{\pi}}

\newcommand{\MD}{\mbox{MD}}
\newcommand{\PAC}{\mbox{PAC}}
\newcommand{\farness}{\mbox{farness}}
\newcommand{\SD}{\mbox{SD}}
\newcommand{\OD}{\mbox{OD}}
\newcommand{\YJl}{h_{\lambda}}

\begin{document}

\def\spacingset#1{\renewcommand{\baselinestretch}%
{#1}\small\normalsize} \spacingset{1}


\newcommand{\blind}{0}

\if0\blind
{
  \title{\bf Class maps for visualizing\\
	           classification results}		
  \author{Jakob Raymaekers, Peter J. 
	  Rousseeuw, and Mia Hubert \vspace{.3cm} \\
		Section of Statistics and Data Science,\\
		Department of Mathematics, KU Leuven, Belgium}
	\date{May 19, 2021}
  \maketitle
} \fi

\if1\blind
{
  \phantom{abc}
  \vskip2.0cm
  \begin{center}
	  {\LARGE\bf Class maps for visualizing\\
	           classification results}
	\vskip1.5cm
	{\large \today} 
	\vskip2cm
	\end{center}
} \fi

\begin{abstract}
Classification is a major tool of statistics and
machine learning.
A classification method first processes a training
set of objects with given classes (labels), with 
the goal of afterward assigning new objects to 
one of these classes.
When running the resulting prediction method on 
the training data or on test data, it can happen 
that an object is predicted to lie in a class that 
differs from its given label.
This is sometimes called label bias, and raises the
question whether the object was mislabeled.
The proposed class map reflects the probability
that an object belongs to an alternative class,
how far it is from the other objects in its given
class, and whether some objects lie far from all 
classes.
The goal is to visualize aspects of the
classification results to obtain insight in the data.
The display is constructed for discriminant analysis,
the k-nearest neighbor classifier, support vector 
machines, logistic regression, and coupling 
pairwise classifications. 
It is illustrated on several benchmark datasets, 
including some about images and texts.

\end{abstract}


\spacingset{1.45} 

\section{Introduction}
\label{sec:intro}
  
Classification is a major tool of statistics and
machine learning.
For an extensive introduction to classification
methods see \cite{Hastie:EOSL}. 
A classification method first processes a training
set of objects with given classes (labels), with 
the goal of afterward assigning new objects to 
one of these classes.
When running the resulting prediction method on 
the training data or on validation data or 
test data, it can 
happen that an object is predicted to lie in a 
class that differs from its given label.
This is sometimes called label bias, and raises the
question whether the object might have been 
mislabeled.
Our goal is to visualize aspects of the data
classification to obtain insight in the data.
For consistency, we will depict the predictions
in the vertical direction throughout.

We start with a simple illustrative example.
The floral bud data originate from 
\cite{Wouters2015} and were kindly provided to
us by Dr. Bart De Ketelaere.
They arose from an experiment in a pear orchard 
with the goal of detecting floral buds in their
environment with branches, bud scales, and 
supports for the trees.
The data contain 550 observations and 6 variables.
There are four classes: `branch' (49 members),
`bud' (363), `scale' (94), and
`support' (44).
The classification in Section 3 yields the 
confusion matrix
\begin{equation*}
\left( 
\begin{array}{rrrr}
  46 &   0 &  0 &  3 \\
   0 & 355 &  2 &  6 \\
	 2 &   0 & 90 &  2 \\
	 6 &   1 &  0 & 37 \\
\end{array}
\right)
\end{equation*}
in which the rows are the given classes and 
the columns are the predicted classes.

\begin{figure}[!ht]
\center
\includegraphics[width = 0.7\textwidth]
  {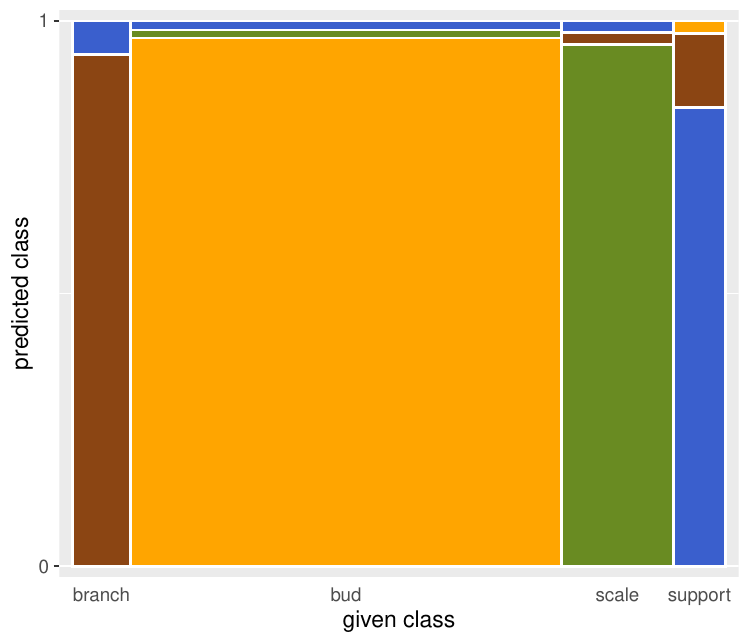}
\vskip-0.2cm
\caption{A stacked mosaic plot of a classification
  of the floral bud data. The 
  given classes (labels) are on the horizontal 
  axis, and the predicted classes are on the 
  vertical axis. The area of each rectangle is
  proportional to the number of objects in it.}
\label{fig:stackplot}
\end{figure}

After a classification is carried out, we can 
display the result in a stacked bar chart or 
a mosaic plot 
\citep{Hartigan1981, Friendly1994}.
Figure \ref{fig:stackplot} shows such a stacked
mosaic plot, which graphically represents the
confusion matrix.
The classes are represented by colors reminiscent 
of their meaning.
The given classes are on the horizontal axis, 
and the predicted labels are on the vertical axis. 
The area of each rectangle is proportional to the 
number of objects in it. The display immediately
shows that the given classes have different 
numbers of objects.
Several variations of this plot are possible. 
One could rank the vertical labels in the order
of the original classes, but we choose to
put the given class at the bottom so that the 
lower part of each bar reflects the objects that
were classified in accordance with their label.
For the other labels in each bar we take
the order of the remaining original classes.
Here we see that buds are often classified 
correctly but that there is some confusion 
between branch and support.  

Figure \ref{fig:stackplot} does not yet give
us an idea why some object is predicted to
belong to a different class.
Is it because the object lies in or near a region 
where classes overlap? Or is it deeply inside 
its predicted class and far from its given class,
arousing suspicion that its original label was 
wrong? Or is it actually far from both its given 
and predicted classes? 
To assist with these questions we will 
propose a display that incorporates additional 
information.

Section \ref{sec:definitions} gives general 
definitions and illustrates the display on an 
example. The subsequent sections apply it 
to discriminant analysis, 
the k-nearest neighbor method, support vector 
machines, logistic regression, and pairwise
coupling.

\section{Basic principles of a class map}
\label{sec:definitions}

Suppose we have objects denoted by their index $i$
where $i=1,\ldots,n$\,, and there are classes
(labels, groups) $g$ with $g = 1,\ldots,G$.
The target is thus a discrete variable with $G$
levels.
Consider a case $i$ in the training set or a test 
set. 
Either explicitly or implicitly, most classifiers
provide posterior probabilities $\hp(i,g)$ of 
object $i$ belonging to each of the classes $g$, 
with $\sum_{g=1}^G{\hp(i,g)} = 1$ for each $i$.
For instance, in discriminant analysis the
posterior probabilities are based on the estimated
densities of the classes, whereas the k-nearest
neighbor classifier estimates them by the
class frequencies in the $k$-neighborhood of $i$.

The object $i$ is then classified according to the
rule
\begin{equation} \label{eq:MAP}
  \mbox{assign object }\; i \; 
	\mbox { to class }\;\;
	\argmax_{g=1,\ldots,G}\; \hp(i,g)\;.
\end{equation}
Now assume that object $i$ has a known given label 
$g_i$\,. 
We wish to measure to what extent the given label 
$g_i$ agrees with the classification.
For this purpose we define the highest $\hp(i,g)$ 
attained by a class {\it different from} $g_i$ as
\begin{equation}\label{eq:altclass}
   \tp(i) = \max\{\hp(i,g)\,;\,g \neq g_i\}\;.
\end{equation}
The class attaining this maximum can be seen as 
the best alternative class. 
If $\hp(i,g_i) > \tp(i)$ it follows
that $g_i$ attains the overall highest value of 
$\hp(i,g)$ so the classifier agrees with the given 
class $g_i$\,. 
On the other hand, if $\hp(i,g_i) < \tp(i)$
the classifier will not assign object $i$ to 
class $g_i$\,. 

We now compute the conditional posterior 
probability of the best alternative class when 
comparing it with the given class $g_i$ as 
\begin{equation}\label{eq:PAC}
	\PAC(i) = \frac{\tp(i)}{\hp(i,g_i) + \tp(i)}\;\;.
\end{equation}
The abbreviation PAC stands for Probability of
the Alternative Class.
Note that when\linebreak 
$\PAC(i) < 0.5$ the classifier does predict the 
given class $g_i$\,, whereas $\PAC(i) > 0.5$ 
indicates that the best alternative class 
outperforms $g_i$ in the eyes of the classifier.
$\PAC(i) \approx 0$ corresponds to a class that fits 
very well, and $\PAC(i) \approx 1$ to a class that 
does not fit well at all.
In cluster analysis, the silhouette plot of
\cite{Rousseeuw:Silhouettes} is based on the same 
idea of quantifying how well an object is positioned
in its class by a comparison with the best
alternative class.

Note that the denominator of \eqref{eq:PAC} is 
strictly positive by construction. When there are 
$G=2$ classes, the alternative class is the only 
other class, the denominator of \eqref{eq:PAC} 
simplifies to 1, and $\PAC(i)$ becomes an 
unconditional probability. 
The $\PAC$ bears some similarity to the notion 
of margin given by $m(i) = \hp(i,g_i) - \tp(i)$, 
whose distribution
is used by \cite{Cai2009} to measure the 
classification capability of an ensemble of 
classifiers.

The second component of the proposed display 
reflects how far the object $i$ is from its given
class $g_i$\,, again in the eyes of the classifier.
For this we start by computing a distance $D(i,g_i)$
of $i$ to $g_i$\,. The choice of $D$ needs to be
relevant for the type of classifier, because 
different classifiers make different assumptions
about the data. For instance, discriminant analysis
assumes roughly elliptical point clouds, k-nearest
neighbors assume no particular shape but focus on
local inter-point dissimilarities, and support
vector machines allow for disconnected classes.
Next we estimate the cumulative distribution 
function (cdf) of $D(x,g_i)$ where $x$ is a random 
object generated from class $g_i$\,. Then we define 
the {\it farness} of the object $i$ to its class 
$g_i$ as that cdf in $D(i,g_i)$, that is,
\begin{equation}\label{eq:farness}
 \mbox{\farness}(i) = P[D(x,g_i) 
                        \leqslant D(i,g_i)]\;.
\end{equation}
Therefore $\farness(i)$ lies in the $[0,1]$ range,
just like $\PAC(i)$.
Our implementation estimates the cdf of $D$ by 
pooling the $D(i,g_i)$ for $i=1,\ldots,n$ 
and applying the function	\texttt{transfo} of the 
\textsf{R}-package \texttt{cellWise} 
\citep{cellWise}. 
This method is described in \cite{TVCN}
and in Section A.1 of the	Supplementary Material. 
	
\begin{figure}[!ht]
\centering
\includegraphics[width = 0.65\textwidth]
  {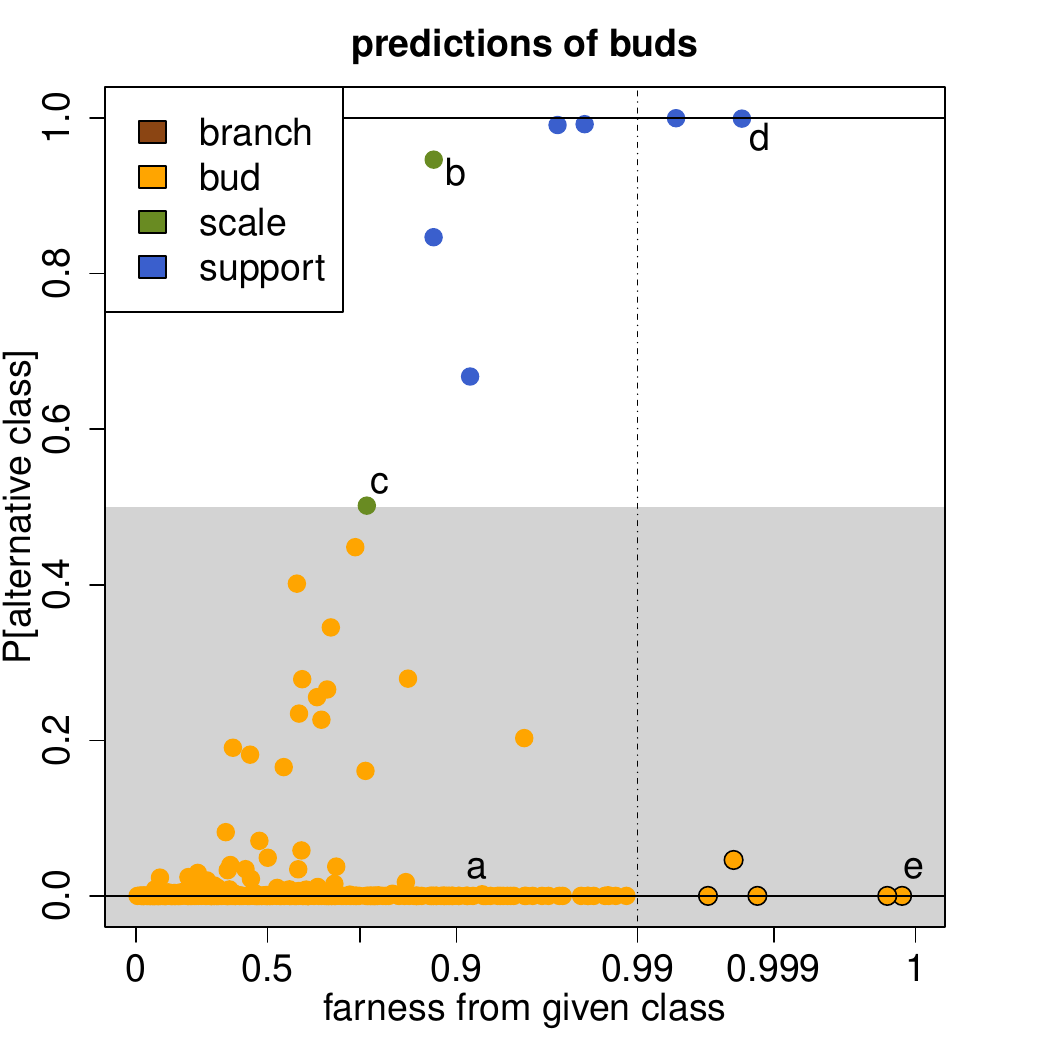}
\caption{Example of a class map.}
\label{fig:classmap_bud}
\end{figure}

We propose to draw a so-called {\it class map} 
for each class $g$, which plots $\PAC(i)$ versus
$\farness(i)$ for all objects $i$ in $g$. 
Figure \ref{fig:classmap_bud} is such a class 
map where $g$ is the class `bud' in 
Figure \ref{fig:stackplot}.
Both axes are expressed in probabilities.
Most points lie in the grey region where 
$\PAC < 0.5$, meaning that the classifier has 
assigned them to their given class, `bud',
and they are shown in orange, the color of
the class `bud' in Figure \ref{fig:stackplot}.
The lower $\PAC(i)$, the more the classifier 
is convinced they belong to their class.
For instance, point \texttt{a} has 
$\PAC(i) \approx 0$.
There are also some points above the grey region, 
where $\PAC > 0.5$ so the classifier has assigned 
them to their best alternative class instead of 
`bud', with much conviction when $\PAC$ is high.
Point \texttt{b} is like that, and its green color 
indicates that it was assigned to the class 
`scale'.
Points with high $\PAC(i)$ do not sit well 
inside their given class, the graphical analogy
being that of a fish out of water (the grey zone).
There can also be boundary cases, such as point 
\texttt{c} with $\PAC \approx 0.5$, meaning
that the classifier does not have much conviction 
to assign it to either its given class or its 
alternative class.

Let us now consider $\farness(i)$ as well.
This is a probability, but the tick marks on 
the horizontal axis are shown on the scale of
the quantiles of the standard Gaussian
distribution restricted to $[0,4]$, in order to
better distinguish the objects with relatively 
high $\farness$.
The dashed vertical line is at 0.99 here, but 
the user can choose a different value if so
desired. (It is not a canonical cutoff like that
of $\PAC(i)$, where $0.5$ is the boundary between 
assigning $i$ to its given class or not.)
Points \texttt{a}, \texttt{b} and \texttt{c}
have typical farness values, meaning they are
relatively close to `bud'. 
We can say that point \texttt{a} is in the best 
quadrant, since it is assigned to its own class
where it is not far from the center.
Point \texttt{b} is not far from the center of
`bud' but nevertheless assigned to `scale'
with much conviction, so there is likely some
overlap between these two classes, with 
\texttt{b} closer to `scale' than to `bud',
and \texttt{c} roughly equally far from both.

What about the remaining two quadrants?
Point \texttt{d} is far from `bud', and at the
same time assigned to `support' (blue color)
with much conviction. In this quadrant the
situation is quite clear: point \texttt{d} 
has two arguments in favor of belonging to its
alternative class `support'.
In the final quadrant, point \texttt{e} lies
quite far from `bud' but it is still 
assigned to `bud' with much conviction.
This can happen when a case is far from its 
class but in the opposite direction of the
best alternative class.

A final piece of information incorporated in the 
class map is whether a point with unusually high 
$\farness(i) = \farness(i,g_i)$ to its own class 
$g_i$ is also far from the other classes. 
To find out we compute the `overall farness' 
of each object $i$ as
\begin{equation}\label{eq:ofarness}
   O(i) = \min_{g=1,\ldots,G} \farness(i,g)\;.
\end{equation} 
When $O(i)$ exceeds the cutoff (here $0.99$) it 
can be considered an outlier, and we put more 
emphasis on that point by using a different 
plotting symbol which has a black border around it. 
For such a point no class seems really suitable.
This happens for point \texttt{e}, but not
for point \texttt{d}.

When we are dealing with training data, the given 
class labels $g_i$ are known, which allows us to 
compute \eqref{eq:altclass}--\eqref{eq:farness}.
For cases without given label, these are not
defined. But we can also have test data or
validation data with given labels. Then we
apply the trained model to the new data,
yielding new posterior probabilities, the
prediction based on~\eqref{eq:MAP}, as well as 
$\PAC(i)$ from \eqref{eq:altclass}--\eqref{eq:PAC}. 
For $\farness(i)$ one needs to
compute the $D(i,g_i)$ with the same parameters
as in the training data, and ensure that 
\eqref{eq:farness}
uses the distribution fitted on the training data.
So for each object $i$ in the test set, 
$\PAC(i)$, $\farness(i)$, and $O(i)$ only depend
on the training set and that case $i$, 
and not on the remainder of the test set.
The test set may even consist of a single case.

The main purpose of class maps is to learn more 
about the objects in a classification.
For instance,
class maps can give an indication of potential 
mislabeling. When cases have a high $\PAC$, and
especially when their $\farness$ is large too,
one may want to check whether their original
label was correct. 
When mislabeling is detected in the 
training data, it may be beneficial to retrain 
the classifier without the mislabeled cases.

Class maps may also reveal subgroups in the data,
which can indicate that a class is not really
homogeneous, and should perhaps be split up.
(An example of a subgroup will be shown
in Figure~\ref{fig:MNIST_classmap_digit1}.)
Class maps may also alert us to the possibility 
that the classifier is less suitable for the data 
at hand.
In Section \ref{sec:comparison} we will 
illustrate this in an example where several 
classifiers are applied to the same dataset.

\section{Discriminant analysis}
\label{sec:DA}

One of the oldest and best understood classification
techniques is discriminant analysis (DA), intended 
for objects that can be represented as points $\bx_i$ 
with $d$ numerical coordinates. 
The underlying model is the normal mixture model as
described in e.g. Chapter~3 of \cite{McLachlan2004}.
It assumes the points in class $g$ follow a normal 
distribution $N(\bmu_g,\bSigma_g)$ with unknown class
mean $\bmu_g$ and covariance matrix $\bSigma_g$\,, and 
unknown class probability $\pi_g$\,.
To train the classifier we compute estimates $\bhmu_g$
and $\bhSigma_g$ which are typically the empirical
mean and covariance of the points $\bx_i$ in class $g$.
The class probability $\pi_g$ is estimated as 
$\hpi_g := n_g/n$
where $n_g$ is the number of objects in class $g$.
The estimated density of the mixture distribution is 
then $\sum_g{\hpi_g \hf_g(\bx)}$ 
with $\hf_g$ the multivariate normal density
\begin{equation}\label{eq:gaussian}
 \hf_g(\bx) =
	\frac{1}{\sqrt{(2\pi)^d \det(\bhSigma_g)}}
	e^{\displaystyle -\MD^2(\bx,\bhmu_g,\bhSigma_g)/2}
\end{equation}
where $\MD^2(\bx,\bhmu_g,\bhSigma_g) := 
(\bx - \bhmu_g)'\bhSigma_g^{-1}(\bx - \bhmu_g)$ is
the squared Mahalanobis distance.

The Quadratic Discriminant Analysis (QDA) model
allows the $\bSigma_g$ to be different.
On the other hand, the Linear Discriminant Analysis 
(LDA) model assumes that all $\bSigma_g$ are equal, 
so one needs a single covariance estimate. 
We can for instance compute $\bhSigma$ from the
pooled data $\btx_i := \bx_i - \bhmu_{g_i}$ for 
$i = 1,\ldots,n$ where $g_i$ is the given label of
object $i$. 
We then set all $\bhSigma_g := \bhSigma$ 
in~\eqref{eq:gaussian}.

In either case, for a given object $\bx_i$ the 
posterior probability of any class $g$ is thus
$$\hp(i,g) = \frac{\hpi_g \hf_g(i)}
 {\sum_{j=1}^G{\hpi_j \hf_j(i)}}$$
so $\sum_{g=1}^G{\hp(i,g)} = 1$, and we assign
$i$ by the maximum a posteriori classification 
rule \eqref{eq:MAP}.
Applying \eqref{eq:altclass} and \eqref{eq:PAC}
yields the conditional probability of the best 
alternative class $\tg_i$\,:
\begin{equation}\label{eq:daPAC}
	\PAC(i) = \frac{\hpi_{\tg_i} \hf_{\tg_i}(i)}
				    {\hpi_{g_i} \hf_{g_i}(i) + 
				    \hpi_{\tg_i} \hf_{\tg_i}(i)}
\end{equation}
in which the denominators of the posterior
probabilities have canceled.

The second ingredient of the class map is the 
$\farness$.
In the setting of discriminant analysis it is
natural to set the distance $D(i,g)$ of object $i$ 
to class $g$ equal to
\begin{equation} \label{eq:MD}
   D(i,g) := \MD(\bx_i,\bhmu_g,\bhSigma_g)
\end{equation}
which is the well-known unsquared Mahalanobis 
distance in \eqref{eq:gaussian} used by the 
classifier.  
We then compute the farness of object $i$ to any 
class $g$ by formula~\eqref{eq:farness}. 

\begin{figure}[!ht]
\center
\vskip0.3cm
\includegraphics[width = 1.0\textwidth]
                {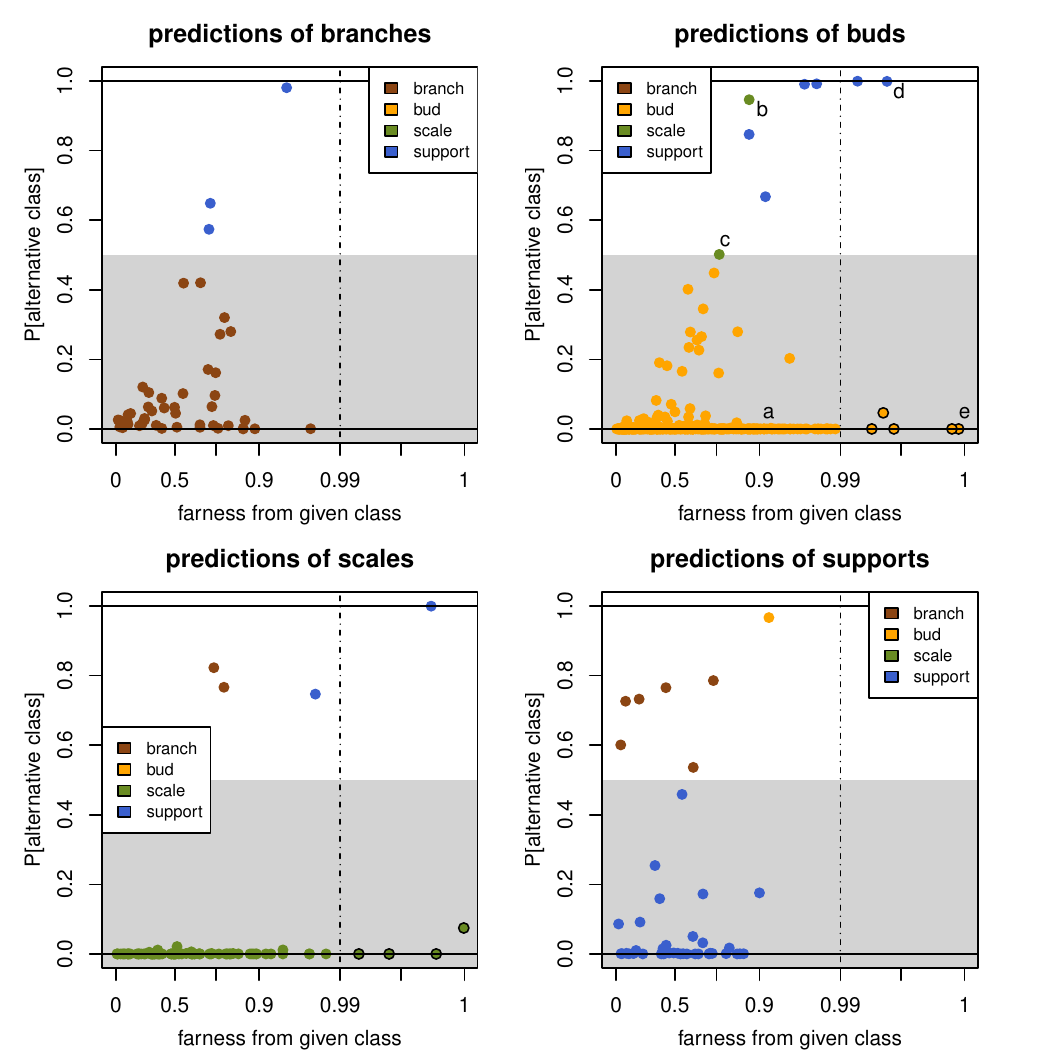}
\vskip-0.2cm
\caption{Floral buds data: maps of all four classes.}
\label{fig:Floralbuds_4plots}
\end{figure}

We classified the floral bud data described in 
the introduction by QDA, yielding the stacked 
plot in Figure~\ref{fig:stackplot} and the class 
map of `bud' in Figure~\ref{fig:classmap_bud}.
Figure~\ref{fig:Floralbuds_4plots} shows the 
class map of each given class separately.
We already interpreted that of buds in 
Section~\ref{sec:definitions}. 
In the class map of scales, most points have a
very low $\PAC \approx 0$, meaning that they
were assigned to their given class with much
conviction.
In the class map of branches we see less
conviction, and the three points with 
$\PAC(i)>0.5$ are in blue so they were assigned 
to class `support'.
Since their farness is unexceptional, they can
be assumed to lie in a region where `branch'
and `support' overlap.
Conversely, in the class map of supports we see
six brown points, so we may conclude that 
`branch' and `support' are not that 
well-separated.

When we have a labeled test set (validation set) 
the same formulas can be applied. When case $i$ 
is in the test set its prediction $\hg_i$ is 
still given by~\eqref{eq:MAP}, and we use the 
formulas~\eqref{eq:daPAC}--\eqref{eq:MD} 
where the $\hpi_g$, $\bhmu_g$, $\bhSigma_g$ and
$\hf_g$ are those of the training set. 


We now illustrate class maps on the MNIST benchmark 
dataset due to \cite{lecun1998gradient}. 
It contains 70,000 images of handwritten digits, 
ranging from 0 to 9, of which 60,000 serve as a 
training set and 10,000 as a test set. Each image has 
been size-normalized and centered in a fixed-size 
image of $28 \times 28$ pixels on a grayscale. 

\begin{figure}[!ht]
\vspace{0.2cm}
\centering
\includegraphics[width = 0.9\textwidth]
                {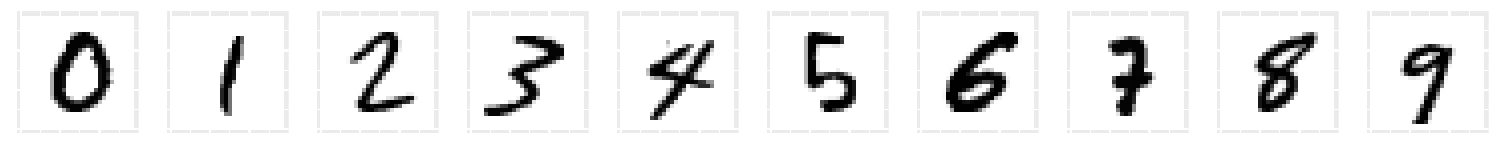}
\includegraphics[width = 0.9\textwidth]
                {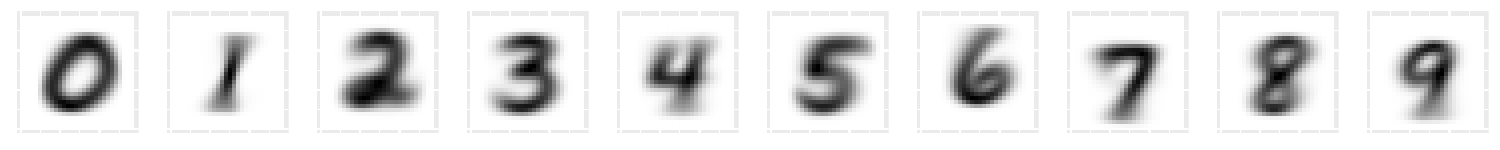}
\caption{Top row: randomly sampled digits;
   bottom row: averaged images per class.}
\label{fig:digit2rows}
\end{figure}

The top row of Figure \ref{fig:digit2rows} shows some
randomly sampled images from the dataset, one of each 
digit, and the bottom row shows the averaged image per 
class.

\begin{figure}[!ht]
\center
\vskip0.2cm
\includegraphics[width = 0.78\textwidth]
  {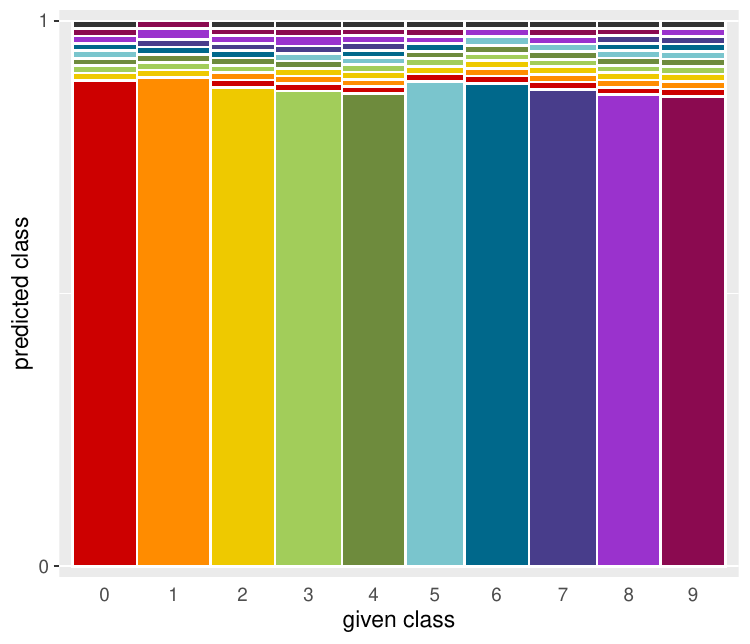}
\vskip-0.2cm
\caption{MNIST data: stacked mosaic plot where the 
objects flagged as outliers are shown in dark grey, 
as an extra class at the top.}
\label{fig:stackplot_MNIST}
\end{figure}

To predict digits from images we first reduced
the dimension of the data from 784 to 50 by PCA.
This avoids numerical instability due to the 
inversion of covariance matrices in QDA.
The misclassification rate of QDA is quite low
here, around 4\%. Figure \ref{fig:stackplot_MNIST} is 
the stacked mosaic plot of the QDA classification. 
Unlike Figure \ref{fig:stackplot},
here we opted to see the outliers [given by 
$O(i) > 0.99$ in \eqref{eq:ofarness}] as an 
extra predicted class in dark grey at the top.
Fortunately there are not many outliers here. 

\begin{figure}[!ht]
\vspace{0.3cm}
\centering
\includegraphics[width = 0.65\textwidth]
  {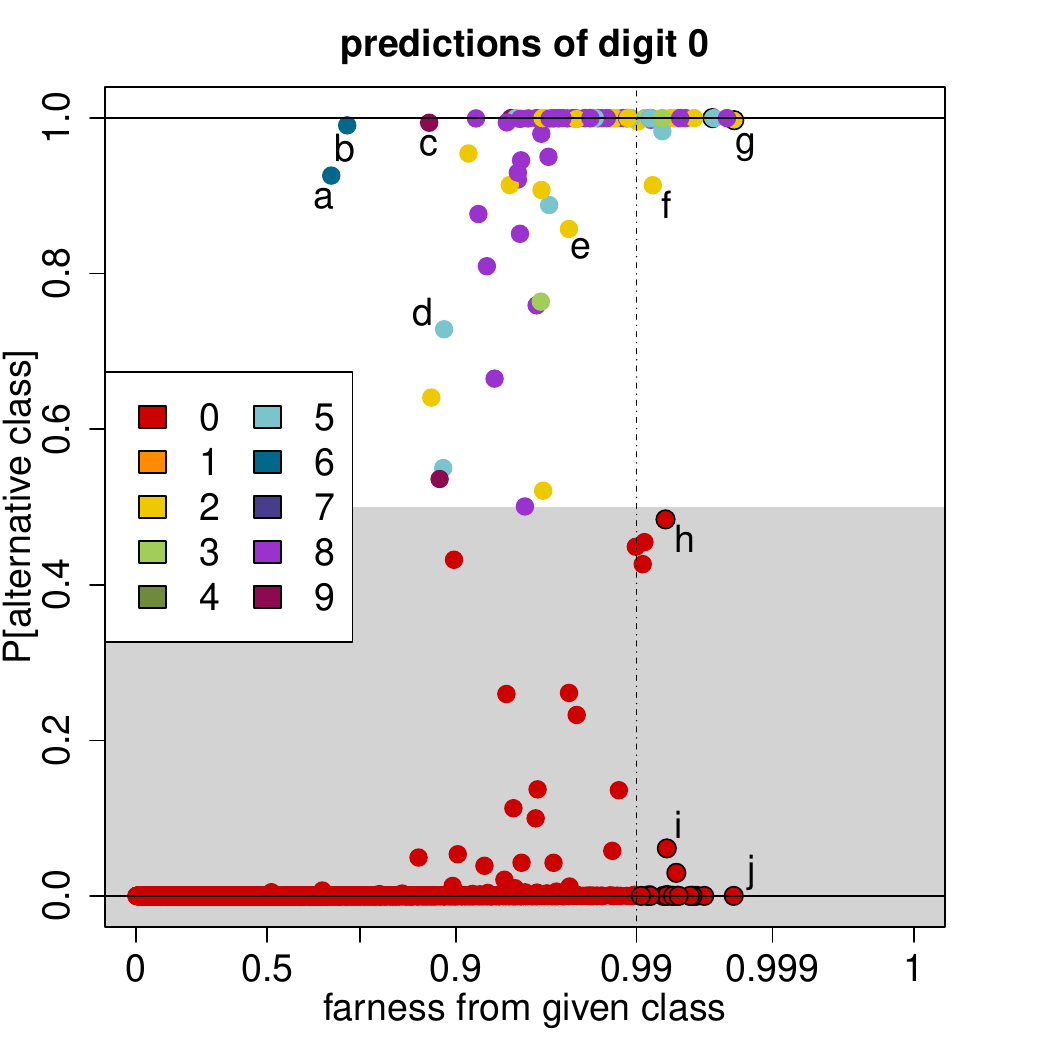}\\
\vspace{0.3cm}
\includegraphics[width = 0.6\textwidth]
  {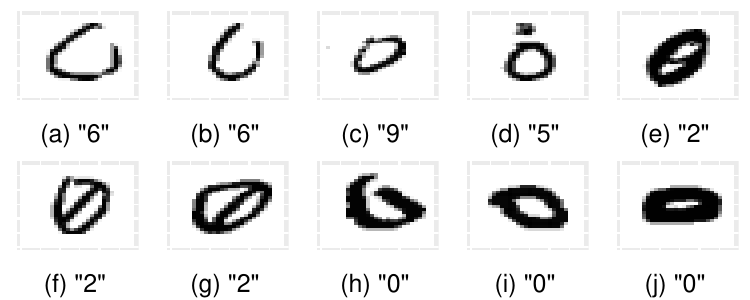}
\caption{Class map of the digit 0, with the images 
  corresponding to the marked points.}
\label{fig:MNIST_classmap_digit0}
\end{figure}

The top panel of 
Figure~\ref{fig:MNIST_classmap_digit0} 
is the class map of the digit 0. 
The large majority of points have $\PAC \approx 0$,
meaning they lie well within the class.
Some more exceptional points have been marked,
and the corresponding images are shown under the 
class map.
Let us start with points above the grey zone, 
that is, whose predictions differ from 0. 
Points \texttt{a} and \texttt{b} have a high $\PAC$ 
but a relatively low $\farness$. We see
that QDA predicts them as a 6, which is not 
surprising when looking at their images. 
We also see why the farness is not very high, since
the images look a lot like a zero, but without 
fully closing the circle.
Point \texttt{c} is a very small circle, positioned
somewhat higher than the average zero. This makes
it look as though it could be the top part of the 
digit 9. QDA indeed considers it a 9, but its low 
$\farness$ indicates that it is not far from 
the class of zeroes.
Point \texttt{d} has a $\PAC > 0.5$, but not quite 
as high as the other marked points. QDA considers
this a 5, and from the image we understand why. 
It is a 0 with an extra horizontal stroke on top,
which gives it several characteristics of a 5. 
Note that its $\farness$ is not extreme, so this
image is in between both classes.
Points \texttt{e} to \texttt{g} are all classified
as 2, and their images are zeroes with an extra line.
In \texttt{f} and \texttt{g} the extra line becomes 
more slanted and pronounced, increasing the 
resemblance with the digit 2 which also has such 
a slanted stroke, and increasing their $\PAC$. 
Note that these two points also have a higher
$\farness$ and that \texttt{g} has a black border 
so it is an outlier, indicating that it is rather 
distant from all classes.
Finally, points \texttt{h}, \texttt{i} and 
\texttt{j} are outliers with a $\PAC$ below 0.5,
so they are assigned to class 0, but their 
images are flat zeroes written with thick pen 
strokes, unlike most zeroes in the data.

\begin{figure}[!ht]
\centering
\includegraphics[width = 0.65\textwidth]
  {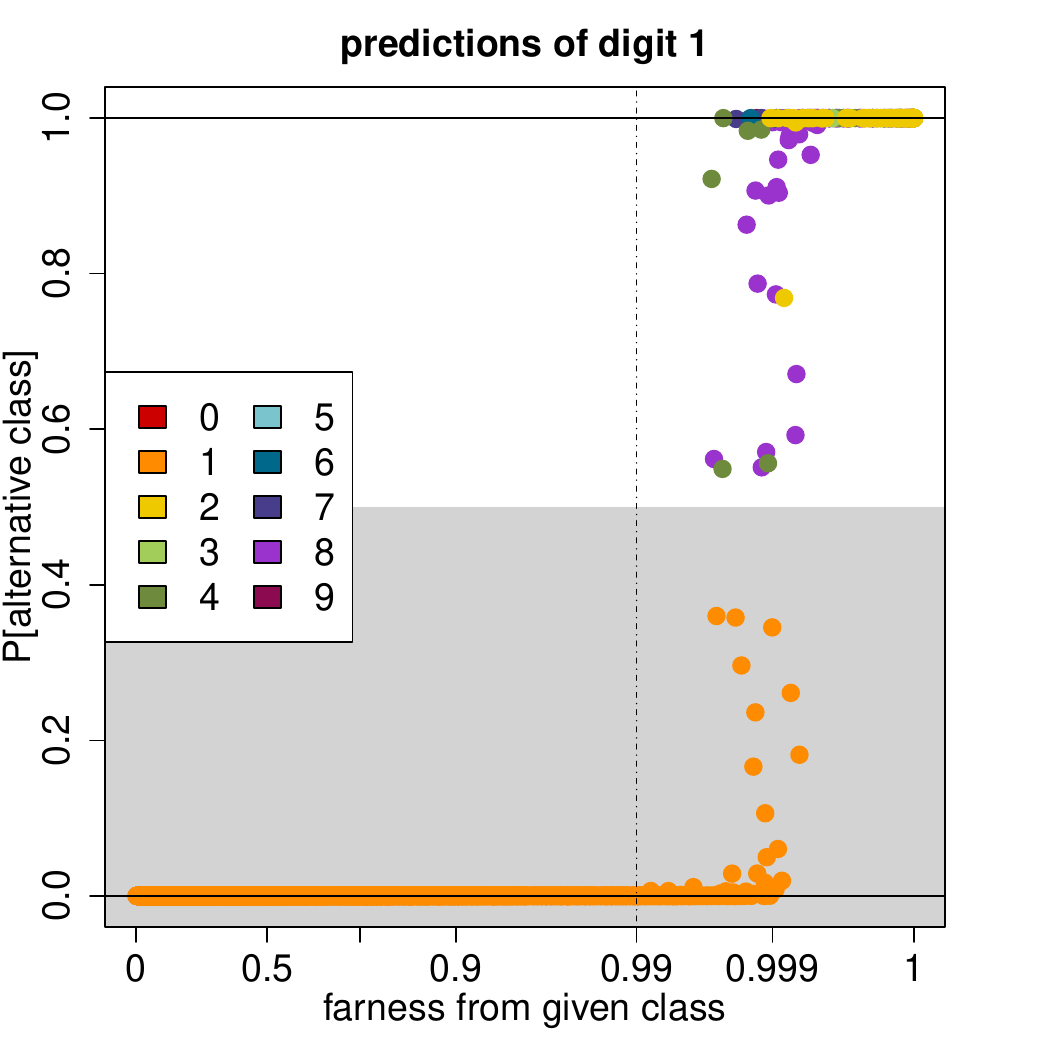}\\
\caption{Class map of digit 1.}
\label{fig:MNIST_classmap_digit1}
\end{figure}

\begin{figure}[!ht]
\center
\vskip0.2cm
\includegraphics[width = \textwidth]
  {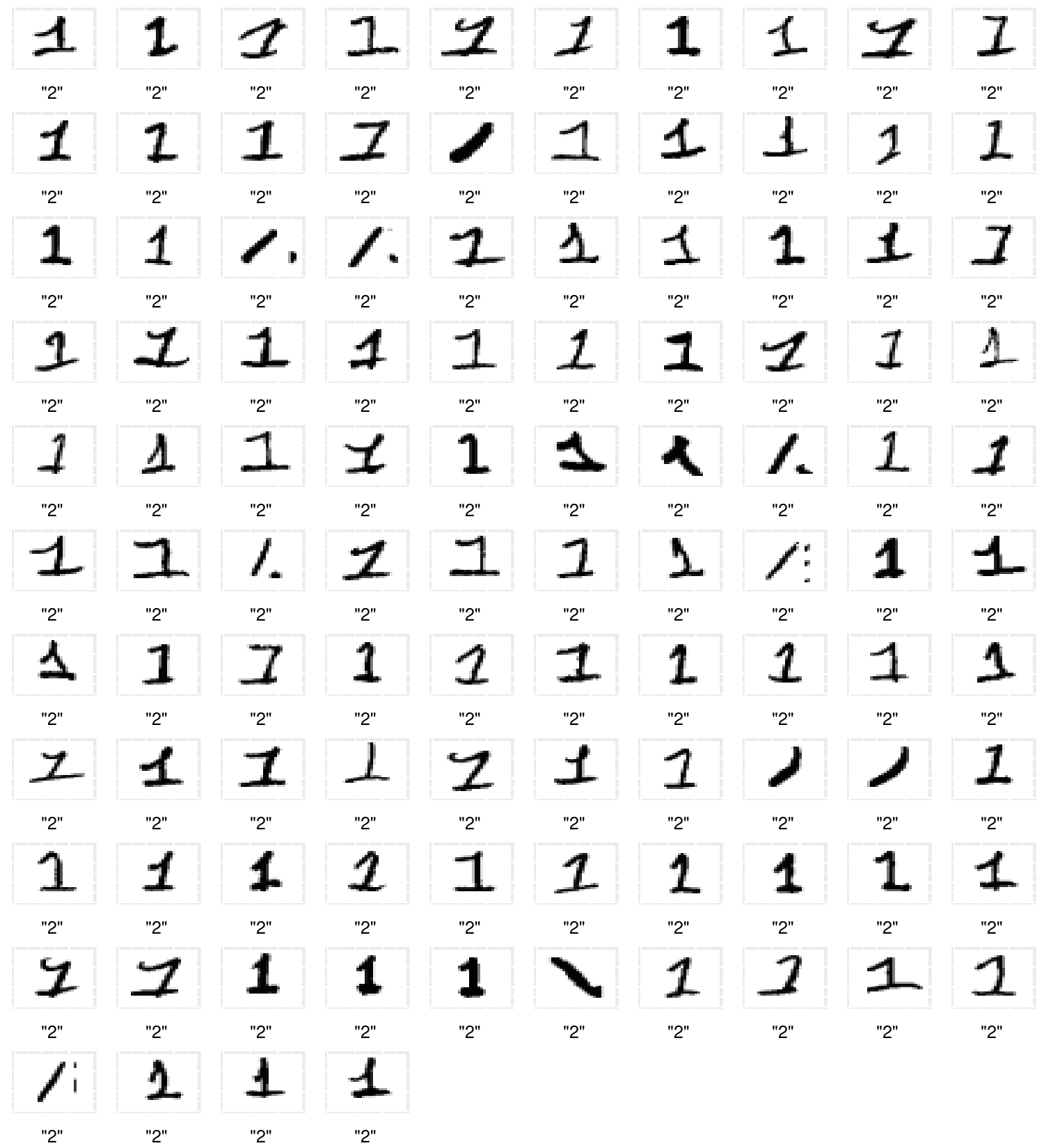}
\vskip-0.2cm
\caption{Images of the digits in class 1 predicted 
   as a 2.}
\label{fig:farnessMNIST_discussion_class1predas2}
\end{figure}

The class map of digit 1 is in 
Figure~\ref{fig:MNIST_classmap_digit1}.
Unlike the other class maps, all the digits
with $\PAC > 0.5$ have a high farness. 
This may be due to the fact that most images
of 1 are just vertical lines, so they look very 
similar to each other.
As a result, deviations from that shape get
a high Mahalanobis distance.
Out of the 6742 images, QDA predicts 306 in a 
different class, so their $\PAC > 0.5$.
Of these, 104 are predicted in class 2 
(yellow points).
The fact that these points are relatively 
concentrated in the class map suggests that they 
may have some properties in common. 
Figure~\ref{fig:farnessMNIST_discussion_class1predas2} 
shows all the images of 1's predicted as 2. 
Most of these digits are indeed 1's but written
with a horizontal line at the bottom (which often
occurs in a 2).
This is unlike the vast majority of 1's in the 
data, as can be inferred from the averaged image
of 1 in Figure~\ref{fig:digit2rows}. 
We have thus identified a subgroup of class 1, which 
corresponds with a particular writing style that 
is not recognized by the classifier.

The class maps of the remaining digits, as 
well as the relevant images, can be found in 
Section~\ref{suppmat:MNIST} of the Supplementary 
Material.

When classifying the MNIST test data using the
trained QDA model, the misclassification rate
is about the same as on the training data.
Therefore the classification is stable, without 
indication that the training data was overfitted.
The class maps of the test data, shown in 
Section~\ref{suppmat:MNIST_test}, also look
similar to those of the training data.

\section{Classification by k-nearest neighbors}
\label{sec:kNN}

Another popular classifier is the {\it k-nearest
neighbor method} (kNN) of \cite{FixHodges:kNN}, 
which has several appealing properties.
It is not restricted to data points with coordinates, 
as it can take data in the form of dissimilarities
$d(i,j)$ between objects.
Such a dissimilarity matrix may for instance 
originate from subjective judgments, in which case
there were no coordinates to begin with, and
the axioms of a metric need not be satisfied.
Of course, if there are coordinates one can always
compute dissimilarities from them. 
This even works when the variables are of mixed 
types. Chapter~1 of \citep{FGID} describes
how one can compute a dissimilarity matrix from
mixed variables of continuous, symmetric binary,
asymmetric binary, nominal and ordinal types,
and this is implemented in the function 
{\texttt daisy()} of the \textsf{R} package 
{\texttt cluster} \citep{cluster_package}.

Around each object $i$ the kNN method determines 
its {\it k-neighborhood} consisting of the
objects $j \neq i$ with the $k$ smallest $d(i,j)$.
Let us denote the $k$-th such dissimilarity as 
$d_i^*$.
Such a neighborhood is not always unique, as 
it can happen that there are other objects $j'$
with the same dissimilarity $d(i,j') = d_i^*$\,.
To make the neighborhood unique a common option is 
to include such points $j'$ as well, so we get 
neighborhoods $N(i)$ with $k(i)$ members where 
always $k(i) \geqslant k$.

The kNN method then estimates the probabilities
$\hp(i,g)$ of object $i$ belonging to each of
the classes $g$ by the relative frequencies
\begin{equation}\label{eq:kNNprob}
   \hp(i,g) := n_i(g)/k(i)
\end{equation}
where $n_i(g)$ counts how many objects in the
neighborhood $N(i)$ have label $g$. These 
estimated probabilities again satisfy 
$\sum_{g=1}^G {\hp(i,g)} = 1$.
The kNN classifier then predicts the label of $i$ 
as the class $\hg_i$ with highest $\hp(i,g)$,
in line with~\eqref{eq:MAP}. 
Also here ties can occur. Some implementations
choose randomly between tied labels.
Our implementation breaks ties by assigning
$i$ to the tied label $g$ for which the average
dissimilarity between $i$ and the members of $g$ 
in $N(i)$ is lowest.

From its definition we see that kNN makes no 
explicit assumptions about underlying distributions, 
and that it can focus on local structure (nearby 
objects) rather than global structure. 
Both aspects are quite different from DA.
For a given dataset, a typical way to select an 
appropriate value of $k$ is to cross validate the
misclassification rate. Here we will assume that
$k$ has already been selected.

The conditional probability of the best 
alternative class $\PAC(i)$ is obtained by 
applying~\eqref{eq:altclass} and~\eqref{eq:PAC}
to~\eqref{eq:kNNprob}.
When $\PAC(i)=0$ it means that all members 
of $N(i)$ have the same label $g_i$\,. 
At the other extreme, $\PAC(i)=1$ says that all
members of $N(i)$ have the same label $\hg_i$ 
which differs from $g_i$\,.
The boundary is again at $\PAC=0.5$, with 
$\PAC(i) > 0.5$ signifying that the predicted 
label $\hg_i$ fits better than the given label 
$g_i$\,. Unlike DA, here $\PAC(i)$ takes discrete 
values, which are multiples of $1/k(i)$.

For computing $\farness$ we can no longer use 
\eqref{eq:MD} since the kNN classifier does not 
require coordinates, only interpoint dissimilarities.
The farness should thus be based on those 
dissimilarities.
For each object $i$ and class $g$ we compute $D(i,g)$ 
as the median of the $k$ smallest dissimilarities 
$d(i,j)$ to all objects $j$ of class $g$. 
For each class $g$ we then divide $D(i,g)$ by
$\median\{D(j,g)\,;\,j\mbox{ belongs to class }g\}$.
This makes the $\farness$ values from all classes
more comparable to each other. (This $D(i,g)$ 
definition did not originate from the literature.)
Finally, we estimate the distribution of the $D(i,g)$
as we did for DA, and compute $\farness(i)$ 
from~\eqref{eq:farness}.
The overall $\farness$ is then given 
by \eqref{eq:ofarness} as before.

The above formulas can also be used when we have a 
test set (validation set) with labels. 
For a test case $i$ we consider a neighborhood $N(i)$ 
consisting only of cases $j$ in the training set. 
Maximizing~\eqref{eq:kNNprob} yields the predicted 
class $\hg_i$ as before.
In the $\farness$ computation we reuse the quantities
$\median\{D(j,g)\,;\,j\mbox{ belongs to class }g\}$
from the training set, as well as the distribution
fitted to the $D(i,g)$ of the training set.

We illustrate the resulting display on the benchmark
{\it spam} data. 
These consist of 4601 emails collected by George
Forman at the Hewlett-Packard Labs, and are labeled as 
spam (1813 mails) or non-spam (2788 mails). 
The data is publicly available in the \textsf{R}-package 
\texttt{kernlab} \citep{kernlab}.
The mails were converted to a numerical characterization 
using 57 variables. Section \ref{suppmat:spam} 
of the Supplementary Material lists the variables and 
their interpretation. 
Unfortunately, the mails themselves are not available.

\begin{figure}[!ht]
\vspace{0.3cm}
\centering
\includegraphics[width = 0.75\textwidth]
                {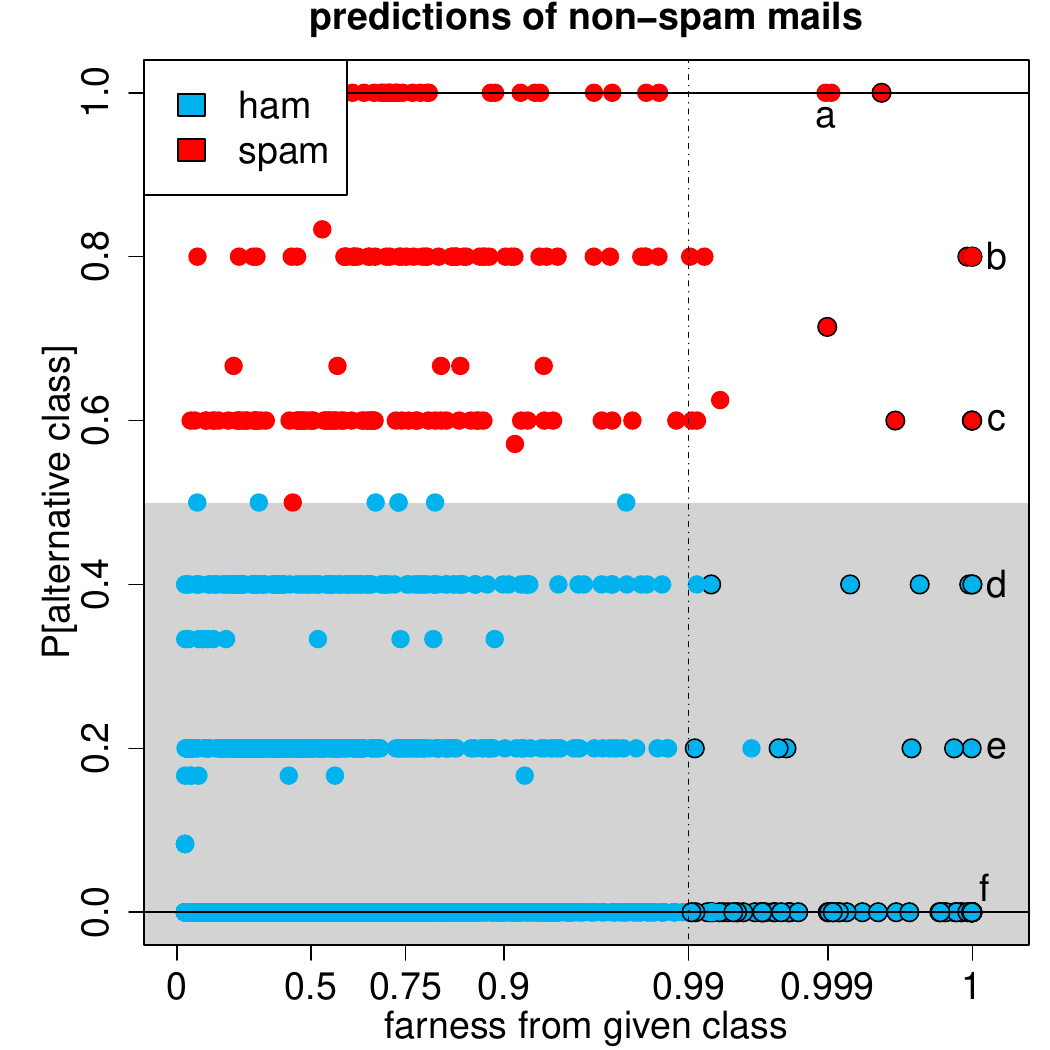}
\caption{Class map of the non-spam mails.}
\label{fig:spam_classmap_ham}
\end{figure}

The data are classified by kNN with $k = 5$,
yielding an in-sample misclassification rate under 9\%.
Figure \ref{fig:spam_classmap_ham} shows the class 
map of the non-spam (also called ham) mails.
As expected, most of the $\PAC$ values are at one of the 
six main levels between $\PAC = 0$ and $\PAC = 1$ by 
steps of $1/k = 0.2$\,. 
The $\PAC$ values in between these levels
come from the neighborhoods $N(i)$ that contain 
$k(i) > k$ members due to tied dissimilarities.

Some atypical points are marked. Point \texttt{a} 
has maximal $\PAC = 1$ so it is strongly predicted 
as spam.
It corresponds to a mail containing 1506 capitalized 
characters, 1488 of which in a single string. 
Capitalization is more common in spam messages, 
explaining why it was predicted as spam. 
Point \texttt{b} is a mail of which 20\% consists 
of the word `free'. 
This word is more common in spam mails, so its
frequent occurrence makes the mail suspicious.  
In mail \texttt{c}, 7.5\% of the characters are `\#'
which appears more often in spam mails than in 
non-spam, and 7.5\% is the highest percentage in 
any non-spam mail. Note that \texttt{b} and 
\texttt{c} are marked as outliers, that is, they 
are unusual and don't lie well within either class.

The next points are correctly predicted as non-spam.
Point \texttt{d} has a high $\farness$ because 30\% of 
its characters are exclamation marks, but it also 
contains some non-spam features such as a high 
frequency of `re' (since spam mails are usually
not replies). 
Mail \texttt{e} has no special characteristics except 
for the highest frequency of the word `report'. 
Mail \texttt{f} is classified perfectly with $\PAC=0$
but has a high $\farness$ because it contains the number 
85 much more often than any other mail in the dataset. 
This number is characteristic for the non-spam class 
however, since it occurs in all the telephone and fax 
numbers of the HP labs, including those of the person
collecting the data. 

\begin{figure}[!ht]
\vspace{0.3cm}
\centering
\includegraphics[width = 0.75\textwidth]
                {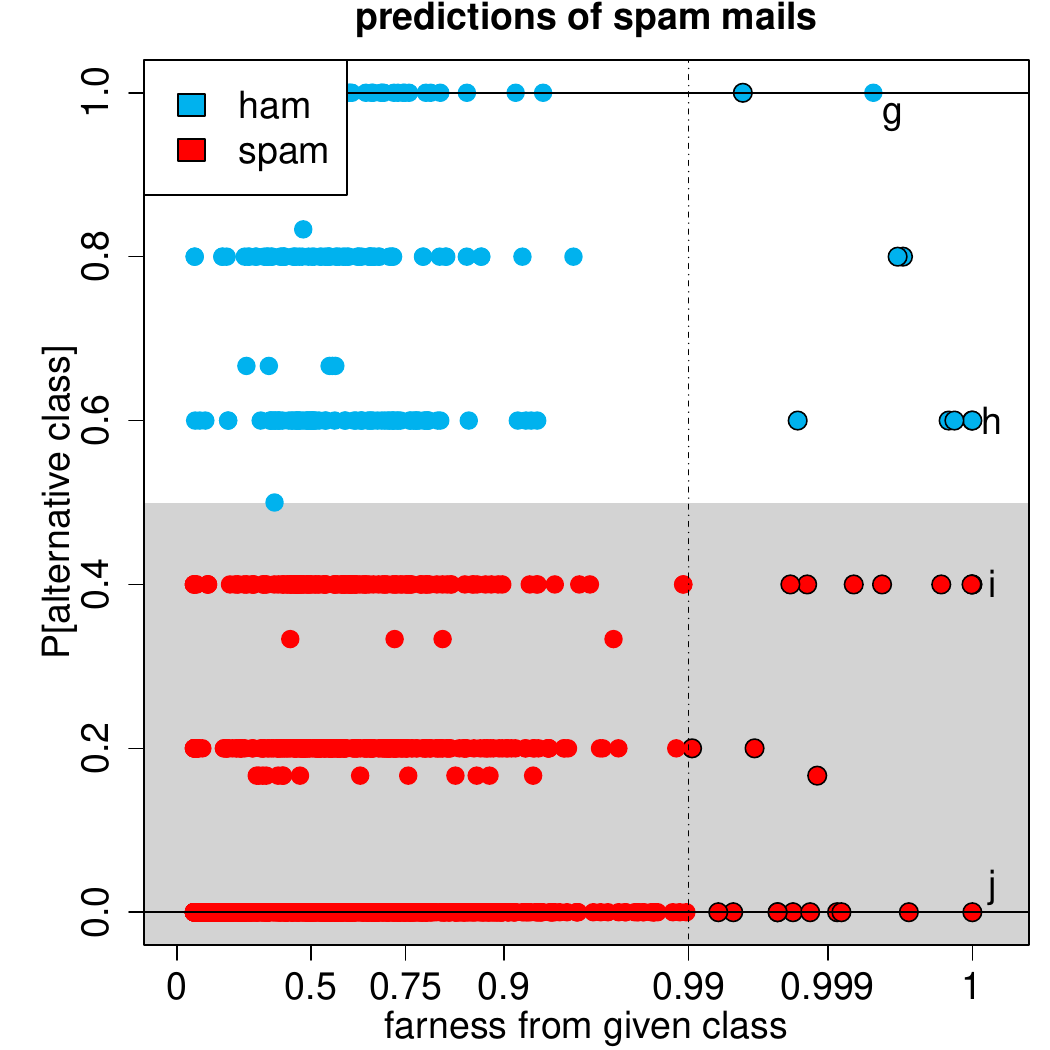}
\caption{Class map of the spam mails.}
\label{fig:spam_classmap_spam}
\end{figure}

Figure \ref{fig:spam_classmap_spam} is the class map 
of the spam messages. Also here some points stand out.
Mail \texttt{g} has $\PAC=1$ so the classifier predicted
it as non-spam with high conviction. It has a very high 
frequency of the round bracket character, typically 
associated with non-spam mails. 
Similarly, mail \texttt{h} also contains many round 
brackets, in addition to the word `technology'.
On the other hand it also has an extreme number of `\#'
symbols (indicative of spam) which explains its high 
$\farness$. 
Mail \texttt{i} contains a string of 9989 capitalized 
characters. This causes it to be correctly classified
as spam, but it also gets a high $\farness$. 
Similarly, mail \texttt{j} is perfectly (since $\PAC=0$)
classified as spam, but has a high $\farness$ as well.
It contains a very high frequency (almost 20\%) of the 
word `credit'. This is a common word in spam messages, 
but 20\% is unusually high.

\section{Support vector machines for two classes}
\label{sec:SVM}

A support vector machine (SVM) is based on a kernel.
Starting from a training set with $n$ objects, the
kernel matrix $K$ is of the form
$\{K(i,j);\;i=1,\ldots,n\mbox{ and }j=1,\ldots,n\}$.
The values $K(i,j)$ play the role of inner products,
unlike the entries of the dissimilarity matrix in
the kNN method which played the role of distances.
The kernel may be derived from a coordinate data 
set $\{\bx_1,\ldots,\bx_n\}$. 
Many kernel functions exist for that situation.
The linear kernel is just 
$K(i,j) = \langle \bx_i,\bx_j \rangle$
where $\langle \, , \, \rangle$ is the usual 
inner product.
The polynomial kernel is
$K(i,j) = (\gamma \langle \bx_i,\bx_j \rangle +
           c)^{\mbox{\footnotesize{degree}}}$
with tuning constants $\gamma$, $c$ and the degree.
The radial basis kernel is given by
$K(i,j) = \exp(-\gamma||\bx_i - \bx_j||^2)$.
Each of these can be seen as 
$K(i,j) = \langle \Phi(\bx_i),\Phi(\bx_j) \rangle$ 
where $\Phi()$ 
maps the original data to a feature space.
For the linear kernel we can just take 
$\Phi(\bx) = \bx$ so there is no transformation.
The other two kernels have a feature space of
a higher dimension than the original space,
in fact for the radial basis kernel that dimension
is even infinite. However, the feature space is
often left implicit since all computations can be
carried out on the kernel matrix $K$ itself.
An advantage of kernels is the added
flexibility, as it is often easier to 
separate classes in a higher dimensional
feature space than in the original space.

The SVM applies the support vector (SV) classifier 
in feature space, i.e. to data $\bv = \Phi(\bx)$.
The SV classifier is a method for $G=2$ classes
which looks for a linear boundary that separates 
the classes as well as possible.
This is achieved by an optimization with a tuning
constant \texttt{cost} that determines to what
extent some points are allowed to be poorly 
classified.
The value of \texttt{cost} is typically selected 
by cross-validation.
The end result is an estimated vector 
$\bhbeta$ and intercept $\hat{\beta_0}$ yielding
the prediction
\begin{equation}\label{eq:SvmRule} 
  \hg(\bv) = 
  \begin{cases}
   \; 1 & \mbox{ if } \quad 
		 \hat{\beta_0} + \langle \bv,\bhbeta \rangle
		  > 0 \\ 
   \; 2 & \mbox{ if } \quad
	   \hat{\beta_0} + \langle \bv,\bhbeta \rangle
		 \leqslant 0 \;.
\end{cases}
\end{equation}
One often calls
$\hat{\beta_0} + \langle \bv,\bhbeta \rangle$
the decision value.
Note that also classification by linear discriminant 
analysis (LDA) is in function of a quantity 
$\hat{\beta_0} + \langle \bv,\bhbeta \rangle$
but it is not the same as it derives from a different 
optimization.

The standard \textsf{C++} library for SVM is LIBSVM 
by \cite{ChangLin2019}, and it is called by the 
function \texttt{svm} in the \textsf{R}-package 
\texttt{e1071} \citep{e1071_package}.
They compute posterior probabilities $\hp(i,1)$ 
by a form of logistic regression with the decision
value as the only regressor, and put
$\hp(i,2) = 1 - \hp(i,1)$. 
The final prediction is then given by~\eqref{eq:MAP}.
The conditional probability of the best alternative 
class $\PAC(i)$ then follows immediately by 
applying~\eqref{eq:altclass} and~\eqref{eq:PAC}.

Farness needs to be defined in relation to how
the classifier works. The SVM attempts to linearly
separate the points in the feature space.
Since the kernel matrix is invariant to 
multiplying the $\bv_i$ by an orthogonal matrix, 
also the $\farness$ needs to be.
Orthogonally equivariant linear structures suggest
using principal component analysis (PCA).
Since the PCA must be applied in the feature space
it is, in fact, kernel PCA (KPCA), but here we will 
write everything in terms of the feature vectors 
$\bv_i$\,.

Carrying out a PCA on the $\bv_i$ in class 1 yields 
an estimated center $\bhmu_1$ of class 1, a matrix
$\mathbf{\hat{U}}_1$ with the 
loadings in its columns, and  scores 
$\bt_i = (\bv_i - \bhmu_1)\mathbf{\hat{U}}_1$
for all $i=1,...,n$\,.
Here we keep all components, so the scores $\bt_i$
have the same dimension as the space $\bV_1$ spanned 
by the $\bv_i$ in class 1.
We then compute the {\it score distance} of any
object $i$ relative to class 1 as
\begin{equation} \label{eq:SD}
  \SD(i,1) = \sqrt{ \sum_j \Big ( 
	  \frac{t_{ij} - \median_h(t_{hj})}
		{\mad_h(t_{hj})} \Big )^2 }
\end{equation}
where $\mad$ is the median absolute deviation.
In this formula $h$ ranges over the members of
class 1, whereas $i$ can belong to either class.
We compute $\SD(i,2)$ in the same way.

When the spaces $\bV_1$ and $\bV_2$ are equal, 
the score distances are all we need. 
Otherwise, $\bV_1$ and/or $\bV_2$ is a proper 
subset of the space $\bV$ spanned by the $\bv_i$ 
of both classes together.
This often happens when using the radial basis 
kernel because then the dimension of $\bV$ can be 
very high (but not infinite).
In such situations a point $\bv_i$ of class 1 may 
not be in $\bV_2$\,.
We then compute how far $\bv_i$ is from $\bV_2$ 
by the euclidean distance between $\bv_i$ and its 
projection on $\bV_2$ given by
\begin{equation} \label{eq:OD}
  \OD(i,2) = ||\bv_i - (\bhmu_2 + (\bv_i - \bhmu_2)
	      \mathbf{\hat{U}}_2 \mathbf{\hat{U}}_2')|| 
\end{equation}
which is called the {\it orthogonal distance},
and $\OD(i,1)$ is computed analogously.

Next we have to combine the score and orthogonal
distances into a single $\farness$ measure.
For this we first scale all $\SD(i,g)$ by the
median of the $\SD(h,g)$ where $h$ ranges
over the members of class $g$.
Next we scale all $\OD(i,g)$ by the median
of the $\OD(h^*,g)$ where $h^*$ ranges
over all cases {\it not} belonging to class $g$
(since $\OD(i,g)=0$ when $i$ belongs to $g$).
Then the distance $D(i,g)$ of an object $i$ to a 
class $g$ is given by
 \begin{equation}\label{eq:SD2OD2} 
  D(i,g) = \sqrt{ SD^2(i,g) + OD^2(i,g) }\;\;. 
\end{equation}
This definition is new, but when the class $g$ 
has a Gaussian distribution
in low dimension, $D(i,g)$ is very similar to
the Mahalanobis distance~\eqref{eq:MD}.
Finally we estimate the distribution of $D$ as in
the previous sections, and 
apply~\eqref{eq:farness} to obtain $\farness(i)$.

Note that SVM can also be applied to $G>2$ classes
by combining pairwise classifications.
Since this technique is also used with other
classifiers, it will be discussed separately in 
Section \ref{seq:pairwise}.

We illustrate the class maps for SVM on a benchmark
data set in which the data do not originate from 
coordinates (measurements).
It is one of the datasets collected and studied by 
\cite{Prettenhofer2010} and consists of 4000 book
reviews on Amazon. 
The reviews were binned into two categories: positive 
(with 4 or more stars out of 5) and negative 
(under 3 stars). 
The 4000 reviews were split up, the first 2000
forming the training set and the next 2000 the 
test set.

The data are actual texts, some of them quite long. 
The kernel matrix was constructed by a string kernel, 
in fact the function {\texttt kernlab::stringdot()} 
with \texttt{type="spectrum"} and \texttt{length=7}.
Afterward the SVM was trained with parameter 
\texttt{cost=2} using the \textsf{R}-package 
\texttt{e1071}. 
The combination of \texttt{length} and \texttt{cost} 
was selected by 10-fold cross validation.

On the training data itself the SVM overfits,
with not a single misclassified book review. 
Perhaps this is not so surprising since 
kernel PCA requires as many as 1803 components to 
explain 95\% of the variance, so we are in a truly
high dimensional setting.
Since the training data has rather low PAC values, 
its class maps are relegated to 
Section \ref{suppmat:bookreviews} 
of the Supplementary Material.
On the test data, the trained SVM obtained
a correct classification rate of 82\%. 

\begin{figure}[!ht]
\vspace{0.3cm}
\centering
\includegraphics[width = 0.75\textwidth]
   {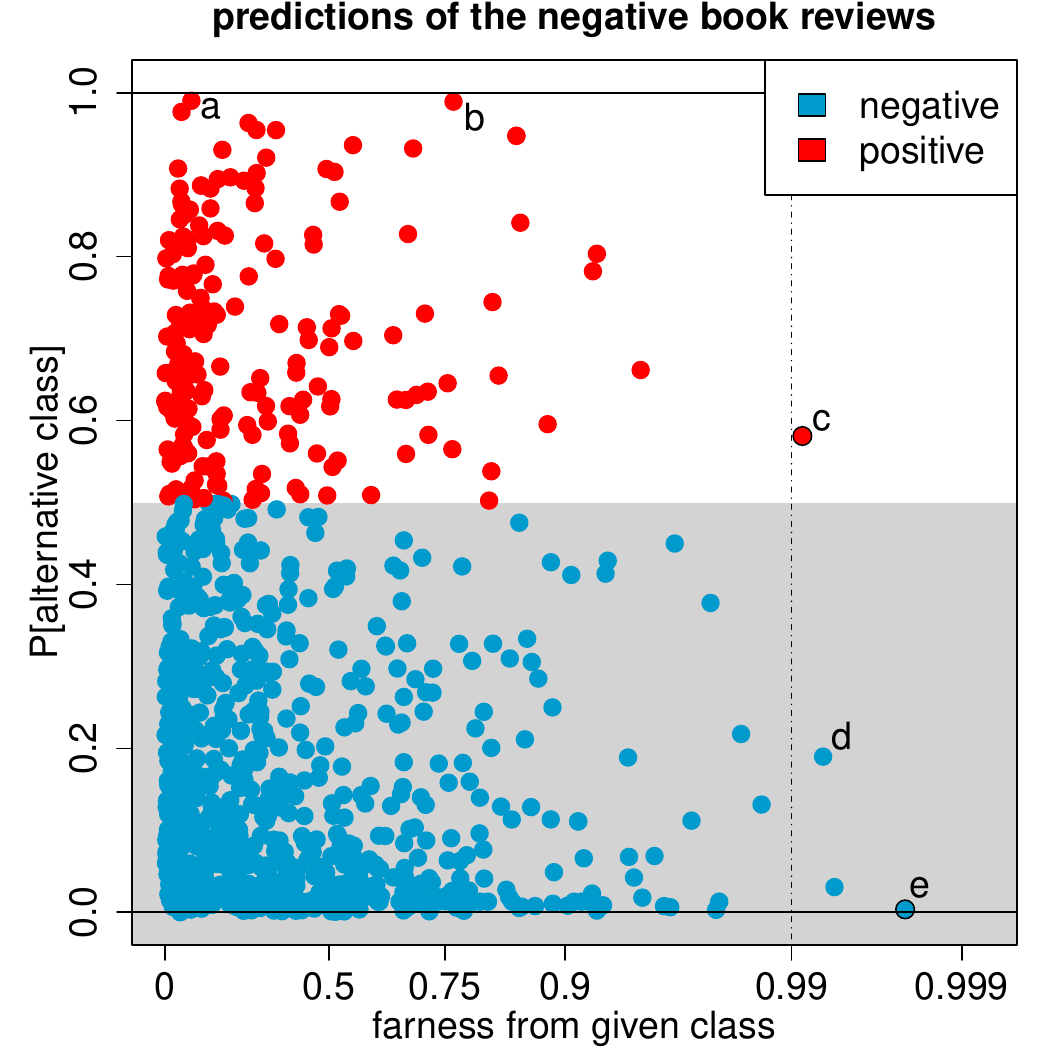}
\caption{Class map of the negative book reviews in 
         the test data.}
\label{fig:bookreview_test_negative}
\end{figure}

\begin{table}[!ht]
\centering
\caption{Excerpts of the negative reviews 
         \texttt{a} to \texttt{e}}
\begin{tabular}{|p{13mm}|p{130mm}|}
\hline
marked & excerpts from the book reviews\\
\hline
\texttt{a} & ``ptt may well be one of heinein's 
    masterworks''\newline 
		``this collection looses the continuity that 
		made the ptt a great read''\\
\texttt{b} & ``i have liked his other books in the 
    past'' \newline 
		``this one didn't have enough insight'' \\
\texttt{c} & ``especially aquinas` arguements'',
``you will definitly not understand''\\
\texttt{d} & ``not at all as interesting as i hoped it would be''\\
\texttt{e} & ``i have read history books more interesting than this book.''\newline
``i thought that it would be an interesting work.''\\
\hline
\end{tabular}
\label{tab:excerpts_test_negative}
\end{table}

The class map of the negative reviews 
in the test data
is shown in 
Figure \ref{fig:bookreview_test_negative}.
About 80\% of these reviews were correctly 
classified as negative (blue). 
Let's look at some points that stand out.
Reviews \texttt{a}, \texttt{b} and \texttt{c} 
have very high $\PAC$. Excerpts of these reviews
 are in Table \ref{tab:excerpts_test_negative}.
Why was it so hard to classify them correctly?
Review \texttt{a} has both positive and negative 
elements. The reviewer is positive about the author 
and the stories, but negative about the selection of 
stories presented in this particular book. 
Review \texttt{b} is similar in that it praises the 
author and his past works, but is negative about 
the current book. 
Point \texttt{c} corresponds with a review containing
some grammatical errors. It also contains a substantial
part that is rather neutral and the negative opinion
in the review is rather indirectly expressed.
Review \texttt{d} is definitely negative, but 
rather short (147 characters), which often leads
to larger $\farness$ values due to few matches of 
the string kernel. 
Finally, review \texttt{e} is
again negative, but relatively short and with a 
substantial neutral/uninformative part.	

\begin{figure}[!ht]
\vspace{0.3cm}
\centering
\includegraphics[width = 0.75\textwidth]
  {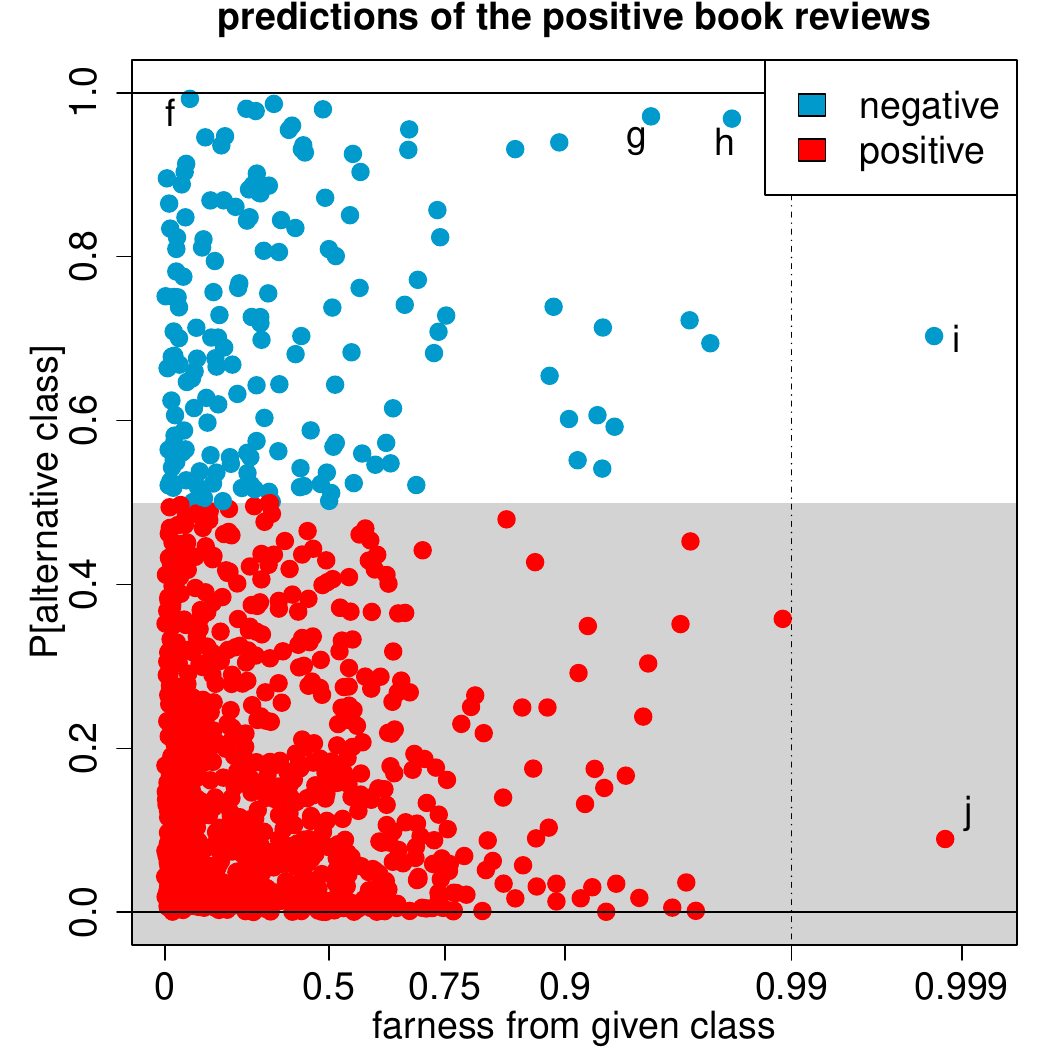}
\caption{Class map of the positive book reviews 
         in the test data.}
\label{fig:bookreview_test_positive}
\end{figure}

The class maps of the positive reviews in the 
test data are in Figure 
\ref{fig:bookreview_test_positive}.
Also here some points stand out from the others,
and we have marked a few for illustration.
Review \texttt{f} has the highest $\PAC$, indicating 
that the classifier strongly wants to put it in 
the negative class. 
It is nevertheless a positive review, but not 
unequivocally so since it also indicates for which 
purposes you should not use this book, as seen
in the excerpt in
Table~\ref{tab:excerpts_test_positive}. 
Book reviews \texttt{g} and \texttt{h} both 
have high $\PAC$ values, and their $\farness$ is 
relatively high. 
These two reviews contain a mix of positive and 
negative comments.
Book reviews \texttt{i} and \texttt{j} have high 
$\farness$ values. This can be explained by the 
fact that these reviews are quite short.
Review \texttt{i} has 101 characters, whereas 
\texttt{j} only contains 22 characters, making 
it the shortest review in the test data. 
The mean length of the reviews is 864 characters.
Review \texttt{i} is assigned to the negative class,
due to a mix of positive and negative comments.

\begin{table}[!ht]
\centering
\vskip0.1cm
\caption{Test data: Excerpts from the positive 
         reviews \texttt{f} to \texttt{j}}
\begin{tabular}{|p{13mm}|p{130mm}|}
\hline
marked & excerpts from the book reviews\\
\hline
\texttt{f} & 
  ``i have not found any other c++ reference that is 
    as complete or as\newline 
	  useful as this books''\newline  
	``this is not an introduction to c++ but rather 
		a reference book''\\
\texttt{g} &
``i felt that the book could have been much better''\newline
``do not expect any groundbreaking plot''\\

\texttt{h} & 
``this book is not for everyone'', 
 ``i was disappointed in the end''\\

\texttt{i} & 
``i really loved the book'', 
`` i just wish the author of the book would''\\

\texttt{j} & 
``like the book very muc''\\
\hline
\end{tabular}
\label{tab:excerpts_test_positive}
\end{table}


\FloatBarrier
\section{Logistic regression}
\label{sec:logistic}

In the context of logistic regression the classes
are typically denoted as 0 and 1 (instead of 1 and
2 as in the other sections), so we are predicting
a binary variable $y$.
The coordinates of the $\bx_i$ can be continuous
and/or binary.
The logistic model assumes that the response in
$\bx_i$ has probability $p(i)$ to be 1 and 
$1 - p(i)$ to be 0, with 
$p(i) = \mbox{logist}(\beta_0 + 
 \langle \bx_i,\bbeta \rangle$).
Here $\mbox{logist}(z) = \exp(z)/(1+\exp(z))$ is 
the logistic function, and the unknown parameters
$\beta_0$ and $\bbeta$ are typically estimated
by maximum likelihood, yielding
$\hbeta_0$ and $\bhbeta$.
Each object then obtains the posterior 
probabilities
\begin{equation} \label{eq:logistic}
  \hp(i,1) = \mbox{logist}(\hbeta_0 + 
  \langle \bx_i,\bhbeta \rangle)
	\;\;\;\; \mbox{ and } \;\;\;\;
	\hp(i,0) = 1 - \hp(i,1)\;.
\end{equation}
Unlike the SVM with two classes, here the 
posterior probabilities follow
directly from the statistical model. 
Formulas \eqref{eq:MAP}--\eqref{eq:PAC}
are applied in exactly the same way, yielding 
the probability of the alternative class 
\begin{equation} \label{eq:logisticPAC}
  \PAC(i) = |y_i - \mbox{logist}(\hbeta_0 + 
  \langle \bx_i,\bhbeta \rangle)|\;\;.
\end{equation}

The $\farness$ measure we use depends on the 
dimensionality of the regressors $\bx_i$\,.
If the regressors are continuous and
low-dimensional we can use the $\farness$ as
in Section \ref{sec:DA} on discriminant analysis,
which is based on the Mahalanobis distance.
We have the choice between estimating a single
$\bhSigma$ for both classes as in linear DA,
or separate $\bhSigma_0$ and $\bhSigma_1$ as
in quadratic DA. 

Note that for multiple {\it linear} regression we 
can replace \eqref{eq:logisticPAC} by the residual 
$y_i - \hy_i$ and plot it versus the Mahalanobis
distance relative to a robust covariance matrix
of the $\bx_i$\,, yielding the outlier map of 
\cite{Rousseeuw:Diagnostic},
later extended to multivariate regression 
\citep{Rousseeuw:MCDreg}.

When the $\bx_i$ are high dimensional it is
less likely that the covariance matrices of
the classes are well-conditioned, and then
one can compute $\farness$ as in 
Section~\ref{sec:SVM} on support 
vector machines. 
Note that in that situation one may prefer to 
run sparse logistic regression, for instance 
using the \textsf{R}-package \texttt{glmnet} 
\citep{glmnet_package}, after which
\eqref{eq:logisticPAC} can be applied.

\section{Combining pairwise classifications}
\label{seq:pairwise}

When the data has $G>2$ labels but the preferred
classifier was designed for 2 labels, like the
support vector machine, one often resorts to 
`one versus one' pairwise classifications.
In this approach one carries out a binary
classification on each pair of classes, yielding
$G(G-1)/2$ comparisons.

These computations yield a matrix with $G(G-1)/2$ 
columns and $n$ rows. 
Each column corresponds to a pair of classes $(g,h)$
with $g<h$ and contains the estimated probabilities
$\hp_{(g,h)}(i,g)$ of the classification into
classes $g$ and $h$.
In the past one typically counted how often each
class came out on top, which is called majority
voting. In recent times the {\it pairwise coupling} 
approach has become more popular.
In this approach one estimates a set of $G$ 
posterior probabilities 
$(\hp(i,1),\ldots,\hp(i,G))$ that match the
pairwise probabilities in the sense that
\begin{equation}
 \hp_{(g,h)}(i,g) \approx \frac{\hp(i,g)}
    {\hp(i,g) + \hp(i,h)}
\end{equation}
for all pairs $(g,h)$. For this one often uses
the second method of \cite{Wu2004} which minimizes
a quadratic loss function.
It is implemented in the function \texttt{couple}
in the \textsf{R}-package \texttt{kernlab}
\citep{kernlab}, with
option \texttt{coupler="minpair"}.

From the resulting posterior probabilities 
$\hp(i,g)$ one immediately obtains the multiclass
prediction from~\eqref{eq:MAP}. Applying 
\eqref{eq:altclass} and \eqref{eq:PAC} 
then yields $\PAC(i)$.

For drawing class maps we also need a measure
of $\farness$. For this we use one of the
methods described in the earlier sections,
a summary table of which is given in
Section A.6. 
If the data have continuous coordinates and
the dimension is not too high, we can compute
the Mahalanobis distance as in \eqref{eq:MD}.
If the data has coordinates of mixed types or
is given in the form of a dissimilarity matrix
we can run the same $\farness$ computation as for 
the $k$-nearest neighbor method in Section
\ref{sec:kNN}.
And if we have high-dimensional continuous data
or the input is a kernel matrix, we can apply
formulas \eqref{eq:SD}--\eqref{eq:SD2OD2}.

As an illustration of pairwise coupling we 
analyze the sweets data. 
This is a subset of the 
\texttt{nutrients\_branded} dataset which is 
publicly available in the \textsf{R}-package 
\texttt{robCompositions} \citep{Templ2020}. 
It contains data on 9 nutritional values of 
804 different sweets sold in Switzerland, which 
are divided into 4 categories: `Cookies and
Biscuits', `Milk based ice cream', `Cakes and 
tarts', and `Creams and puddings'. 
The nutritional variables are the contents of
energy (kcal), protein, water, carbohydrates, 
sugars, dietary fibers, total fat, saturated 
fatty acids, and salt.

\begin{figure}[!ht]
\vspace{0.3cm}
\centering
\includegraphics[width = 1.0\textwidth]
                {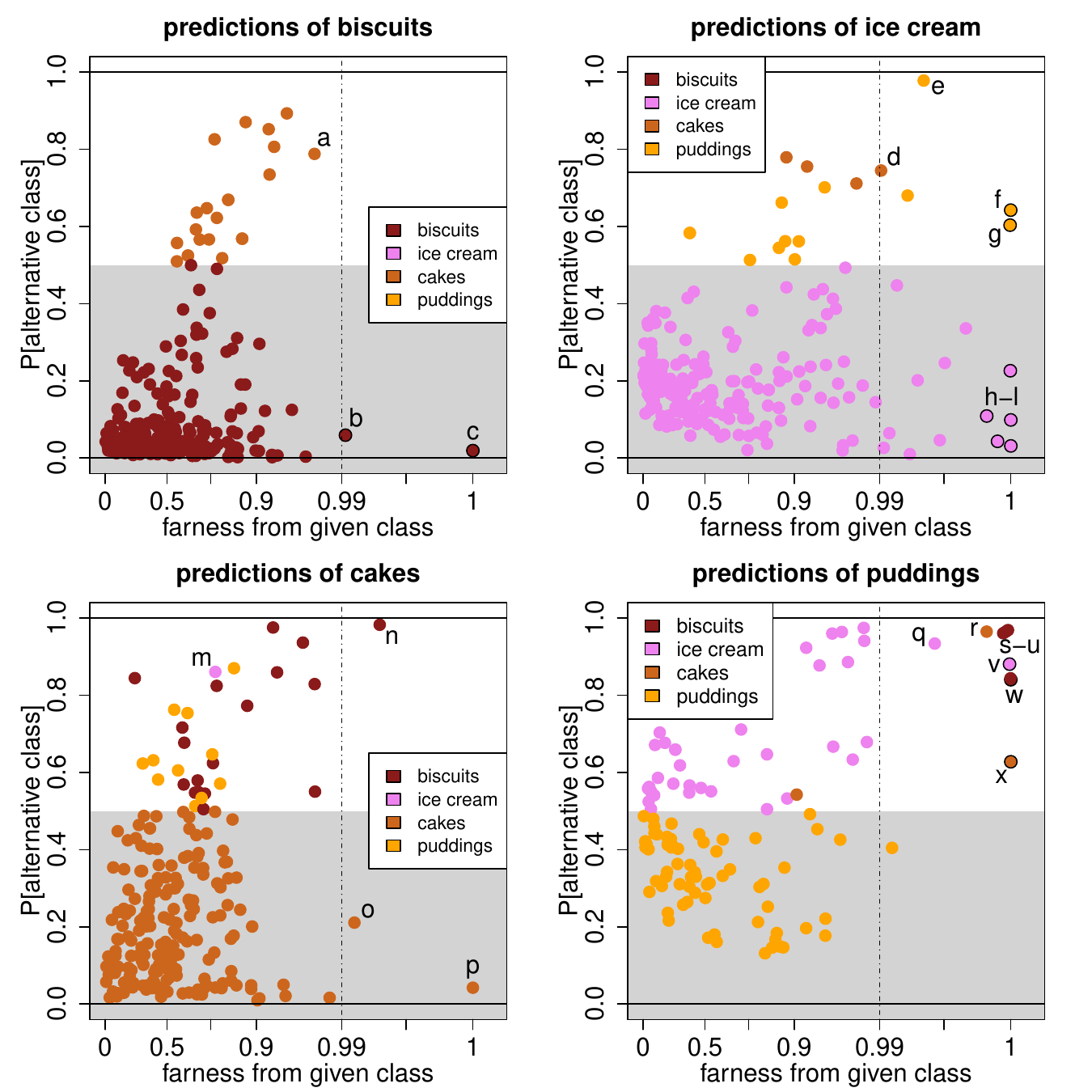}
\caption{Class maps of all four classes in the
         sweets data.}
\label{fig:sweets_4plots}
\end{figure}

We fit an SVM model by the function 
\texttt{svm()} in the \textsf{R}-package 
\texttt{e1071} \citep{e1071_package},
with \texttt{kernel="linear"} and
\texttt{probability=TRUE}. 
The parameter \texttt{cost=10} was selected
by 10-fold cross-validation, yielding a 
misclassification rate of 13\%. 
The function \texttt{predict.svm} with
again option \texttt{probability=TRUE}
internally carries out pairwise coupling 
by the same algorithm of \cite{Wu2004}.
We used the $\farness$ measure of 
Section \ref{sec:SVM} on SVM.
Figure \ref{fig:sweets_4plots} shows the
maps of all classes, whose names
were abbreviated to `biscuits', `ice cream',
`cakes', and `puddings'.

\begin{figure}[!ht]
\center
\vskip0.2cm
\includegraphics[width = 0.7\textwidth]
  {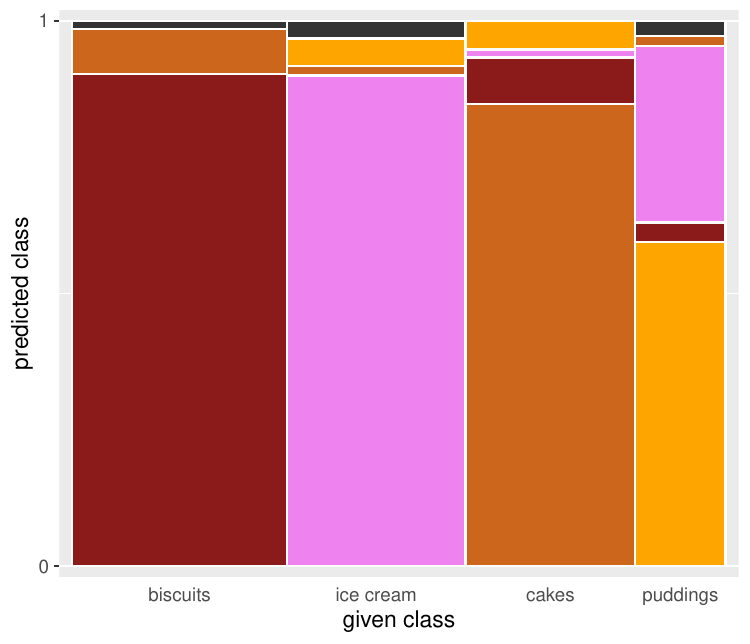}
\vskip-0.2cm
\caption{Sweets data: Stacked mosaic plot of 
the classification. When they occur, the 
products flagged as outliers are shown in dark 
grey, as an extra class at the top.}
\label{fig:sweets_stack}
\end{figure}

Figure \ref{fig:sweets_stack} shows the
stacked mosaic plot of this classification.
We see at a glance that puddings (the yellow
stack at the right) are often predicted as 
pink, the color of the stack of ice cream.
Unlike Figure \ref{fig:stackplot}, here the option 
to show the outliers is switched on. 
They are visible as an extra predicted class in 
dark grey at the top for those classes where 
they occur.

The class maps in Figure \ref{fig:sweets_4plots} 
contain a lot of structure. 
Not being experts in nutrition, we have 
only marked the atypical points we can say 
something about.
The names of the objects \texttt{a} to \texttt{x}
are listed in Section \ref{suppmat:sweets} of the
Supplementary Material.

Let's start with the biscuits. 
A number of points have a $\PAC$ over 0.5, and 
their color indicates they are classified as 
cakes. Most of these are
in fact doughs rather than finished cookies.
They all have a significantly higher water 
content than most of the biscuits in the data. 
This results in them being classified as cakes, 
as cakes are similar to cookies but with a 
higher water content.
The point marked \texttt{a} is a
brownie produced by Burger King. 
Point \texttt{b} has an extreme level of dietary
fibers, actually the highest in the whole data set.
This causes it to be flagged as an outlier, as it
does not really fit well within any of the
classes. Point \texttt{c} is even further 
away. It corresponds to a biscuit with relatively 
high salt and very low water content.

Among the ice creams, the products with 
$\PAC>0.5$ are of two types: one is classified 
as puddings, the other as cakes.
The points classified as cakes correspond to
ice creams that contain a substantial percentage 
of cookie. Point \texttt{d} is a mini
ice cream cone with relatively little ice 
cream compared with the amount of cookie.
The ice creams classified as puddings are nearly 
all Burger King products, with a much higher 
salt content than average for ice cream.
Point \texttt{e} is an example of
such a product, a strawberry sundae. 
There are also several outliers, as indicated
by their black border. Products
\texttt{f} and \texttt{g} have $\PAC>0.5$ 
and are predicted as puddings.
Products \texttt{h} tot \texttt{l} are classified
correctly as ice cream. It turns out that all 
of these outliers, from \texttt{f} to \texttt{l},  
are dietary products.
They have very low sugar, very low fat and high
dietary fiber contents. The points \texttt{f} 
and \texttt{g} also have unusually high salt 
levels, making them more similar to puddings.

In the class map of cakes, object \texttt{m} is 
a strawberry tart, with high water content and 
low sugar relative to most of the cakes.
Due to its high water content it is misclassified
as ice cream.
There is also a group of cakes classified as 
biscuits. 
Nearly all of them have very low water contents,
that are more common for biscuits than for cakes. 
These points correspond to either cake mixes or
fairly dry cakes based on nuts.
The carrot cake \texttt{n} is an extreme example 
of such a product, as it also has very high 
levels of protein and dietary fibers.
The lentil tart \texttt{o} is a dietary product.
It has many characteristics of a cake, but its 
salt levels are rather low and its dietary fiber
is very high. Finally, point \texttt{p}
is a dough, rather than a baked cake. 
 
In the class of puddings the $\PAC$ values are
higher on average than in the other classes, so
this class is the least well-separated.
Most of the misclassified puddings were assigned
to the ice cream class. 
The classifier has a hard time distinguishing 
between pudding and ice cream.
(There is no variable indicating the product's 
storage temperature.)
Most of the puddings misclassified as ice cream
are milk-based, causing relatively high fat 
levels. 
For instance, product \texttt{q} is a chocolate 
mousse.
Point \texttt{r} has both a high $\PAC$ and a
high $\farness$, and is classified as a cake.
Nearly all of its nutritional values are 
exceptional for pudding but normal for cake. 
On the store website it looks like a small 
chocolate cake, so it may actually have been 
mislabeled. 
The ``puddings'' marked \texttt{s} to \texttt{u} 
are in fact dry pudding mixes rather than finished 
products. They have extremely low water and 
extremely high carbohydrate contents, causing
them to be misclassified as biscuits, which
are drier than puddings. 
Point \texttt{v} is an outlier, which leans 
most towards ice cream.
It is a dietary product, with extremely low
sugars and extremely high dietary fibers.
The outlier \texttt{w} is again a dry pudding mix, 
this time with the highest sugar content in the 
whole data set.
Finally, pudding \texttt{x} has the highest salt 
level of any product in the data set. This causes 
it to be flagged as an outlier and classified 
as a cake.


\section{Comparing different classifiers on the 
         same data}
\label{sec:comparison}

\begin{figure}[!ht]
\centering
\includegraphics[width = 1.0\textwidth]
                {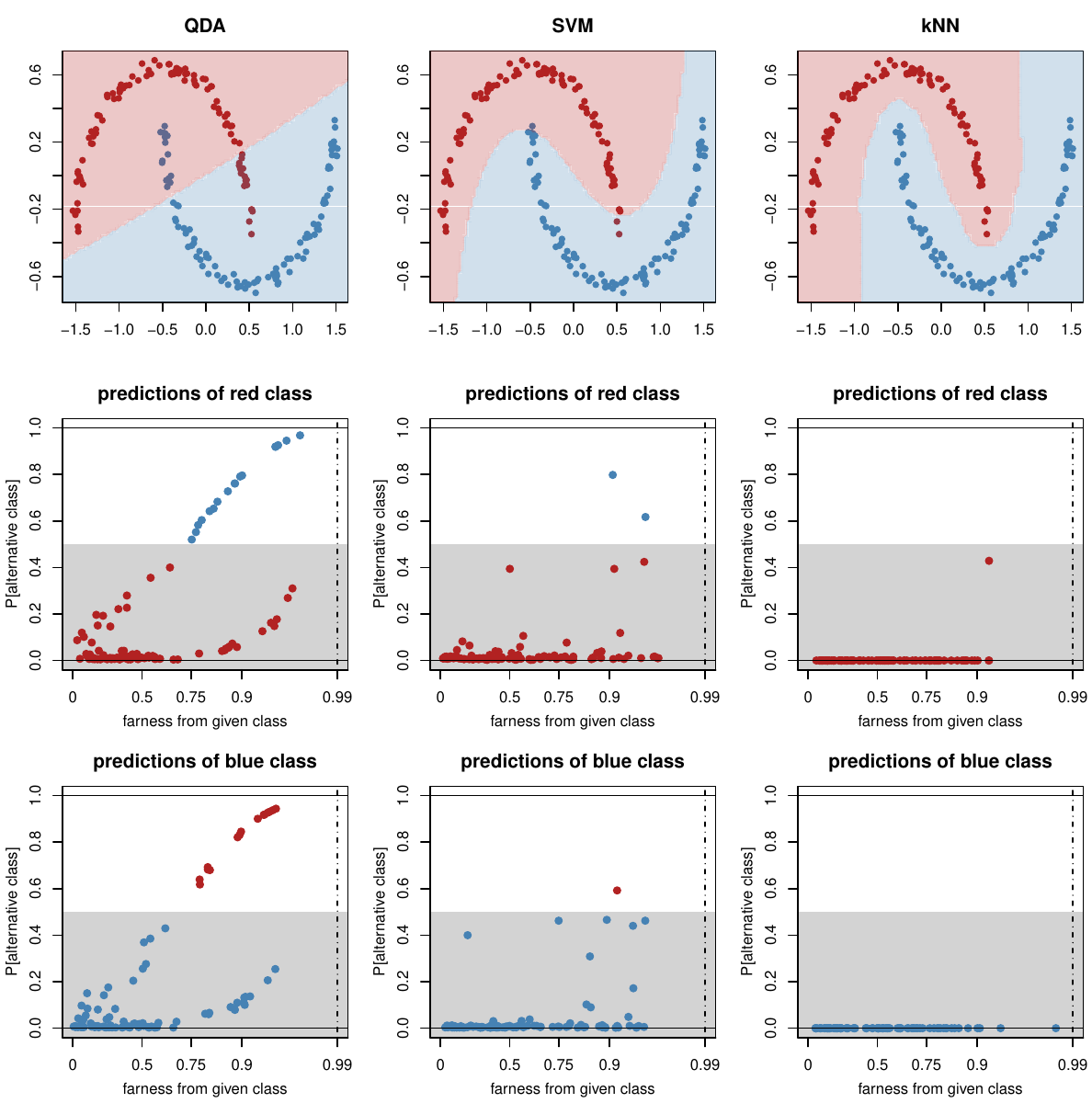}
\caption{Example where three classifiers are
         applied to the same dataset.
				 Each column shows the corresponding
				 classification regions, followed by the 
				 maps of both classes.}
\label{fig:halfcircles}
\end{figure}

Even though class maps are primarily intended to
learn more about the objects in a classification,
it can be interesting to compare the class maps
obtained by several classifiers on the same data.
To illustrate this we consider a simple toy
example with 200 points in two dimensions, for 
which the ground truth is clear. 
The data are shown in the top row of 
Figure~\ref{fig:halfcircles}.
The red and blue classes cannot be separated
by a straight line

The left column of Figure~\ref{fig:halfcircles}
shows the result of the QDA classifier.
The light red and light blue regions in the top
plot were obtained by classifying a dense grid
of new data by the trained model, so a new data
point in the light blue region will be assigned
to the blue class. 
There are about 16\% of misclassifications,
indicating that QDA is not a suitable classifier
here. Its underlying assumption of roughly 
elliptical classes indeed does not hold.
Below this plot is the class map of the red
class. There we see that quite a few points
have $\PAC>0.5$ and are assigned to the blue
class. The class map has a pattern with thin 
threads that correspond to the shape of the 
class. The map of the blue class is similar.

Since the data cannot be separated linearly
in the original space, a kernel transform
may be helpful. Here we applied the radial
kernel with \texttt{gamma = 0.5}, followed by 
the SVM with \texttt{cost = 1.0}. For the
purpose of this illustration we did not 
try to overfit the values of \texttt{gamma} 
and \texttt{cost}.
The results are in the middle column of
Figure~\ref{fig:halfcircles}. 
The assignment regions now have very 
different shapes, and are much better adapted
to the particular forms of these classes.
Now only three points are misclassified, and
they are clearly visible in the class maps. 
A few other points, that lie
close to the boundary between the light blue
and light red regions, have substantial 
$\PAC(i)$ values below $0.5$.

The third column in Figure~\ref{fig:halfcircles}
shows the result of k-nearest neighbors with 
$k=7$. This classifier turns out to work 
best in this particular example. 
It has no misclassifications, so one would likely 
choose kNN here, but the class maps give a finer
picture. Not only are there no points with
$\PAC(i) > 0.5$ (which would be misclassified),
but also below $0.5$ the map looks quite 
different from that of SVM.
Here only a single point has $\PAC(i) > 0$, 
meaning that its k-neighborhood contains points 
from the other class.
Since it occurs in the map of the red class, it
is clear which point it is.

\section{Conclusions}

The proposed class map reflects two basic 
notions, for each object in the class.
In the vertical direction it shows the posterior 
probability of it belonging to its best
alternative class rather than its given class.
In the horizontal direction we see how
far the object is from its given class, with
a separate plotting symbol when it is far
from all classes.

This visualization often provides useful
information about the data, as illustrated 
with examples of discriminant analysis, 
k-nearest neighbors, support vector machines, 
and pairwise coupling.
It also allows to compare the results
obtained by different classifiers.\\

\noindent \textbf{Software availability.}
The methods in this paper have been implemented
in the \textsf{R} package \texttt{classmap} 
\citep{classmap}.
It contains three vignettes that reproduce
the examples shown here.\\

\noindent \textbf{Acknowledgement.} 
This research	was funded by projects of 
Internal Funds KU Leuven. We thank the
Editor, Associate Editor, and two reviewers
for their constructive comments.

%


\clearpage
\pagenumbering{arabic}
\appendix
\numberwithin{equation}{section} 
\section{Supplementary Material} \label{sec:A}
\renewcommand{\theequation}
   {\thesection.\arabic{equation}}

\subsection{Fitting a cdf to distances}
\label{suppmat:transfo}

The definition of farness~\eqref{eq:farness}
requires an estimated cumulative distribution
function of the distance $D(y,g)$ where $y$ 
is a random object generated from class $g$\,.
The available data are the $D(i,g_i)$ of each 
object $i$ to its given class $g_i$\,.
In view of possible heteroskedasticity between
classes, we start by normalizing per class.
For a given class $g$ we divide all the
$D(i,g)$ where $i$ is a member of class $g$ by 
$\median\{D(j,g)\,;\,j 
 \mbox{ belongs to class } g\}$.
The resulting distances are more homoskedastic,
and we pool them to obtain distances
$d_i$ for $i=1,\ldots,n$.
The empirical distribution of the $d_i$ is
typically right-skewed.

In order to account for skewness, we apply
the function \texttt{transfo} of the 
\textsf{R}-package \texttt{cellWise} 
\citep{cellWise} with default options.
This function first standardizes the $d_i$ to
\begin{equation*}
  x_i = \frac{d_i - \mbox{Med}}{\mbox{Mad}}
\end{equation*}
where $\mbox{Med}=\median_{j=1}^n d_i$
and $\mbox{Mad}$ is the median absolute 
deviation given by 
$\mbox{Mad}=1.4826 
 \median_{j=1}^n |d_i-\mbox{Med}|$
as implemented in the standard function 
\texttt{mad()} in \textsf{R}.
Next, \texttt{transfo} carries out the
transform of \cite{Yeo2000} given by
\begin{equation}
\YJl(x) = 
\begin{cases}
((1+x)^{\lambda} - 1) / \lambda &\mbox{ if } \lambda 
    \neq 0 \mbox{ and } x \geqslant 0\\
\log(1+x) &\mbox{ if } \lambda = 0 \mbox{ and } 
    x \geqslant 0\\
-((1 - x)^{2 - \lambda} - 1)/(2 - \lambda) &\mbox{ if }
    \lambda \neq 2 \mbox{ and } x < 0\\
-\log(1 - x) &\mbox{ if } \lambda = 2 \mbox{ and } x < 0
\end{cases}
\end{equation}
which aims to bring the distribution close
to a normal distribution.
The transformation $\YJl$
is characterized by a parameter $\lambda$
that has to be estimated from the data. 
This estimation is typically done by maximum 
likelihood, but the default in \texttt{transfo} 
is to apply the weighted maximum likelihood 
estimator of \cite{TVCN} which is less 
sensitive to outliers.
The resulting $\YJl(x_i)$ are in turn standardized
by their own $\mbox{Med}$ and $\mbox{Mad}$, 
yielding $z_i$ whose distribution is 
approximately standard normal. 
The estimated cdf of the distances $d_i$ is then 
given by $\hat{F}(d_i) := \Phi(z_i)$ where $\Phi$ 
is the standard normal cdf.

\subsection{More on the MNIST data}
\label{suppmat:MNIST}

\begin{figure}[!ht]
\center
\vskip0.5cm
\includegraphics[width = 0.7\textwidth]
              {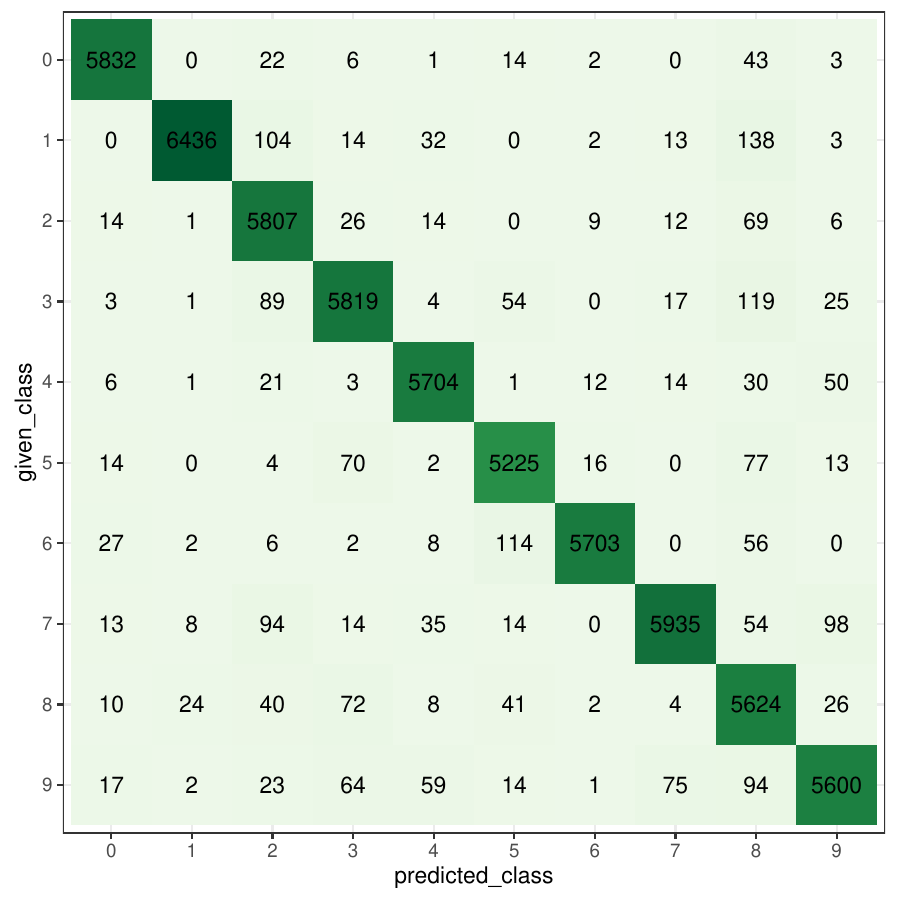}
\caption{Digits data: image plot of the
         confusion matrix.}
\label{fig:imageplot}
\end{figure}

Figure \ref{fig:imageplot} is the image plot of
the confusion matrix of the MNIST data.
Like the stacked mosaic plot in 
Figure~\ref{fig:stackplot_MNIST}, it clearly 
indicates that most cases are assigned to their 
given class.

\begin{figure}[!ht]
\center
\vskip0.2cm
\includegraphics[width = 0.7\textwidth]
                {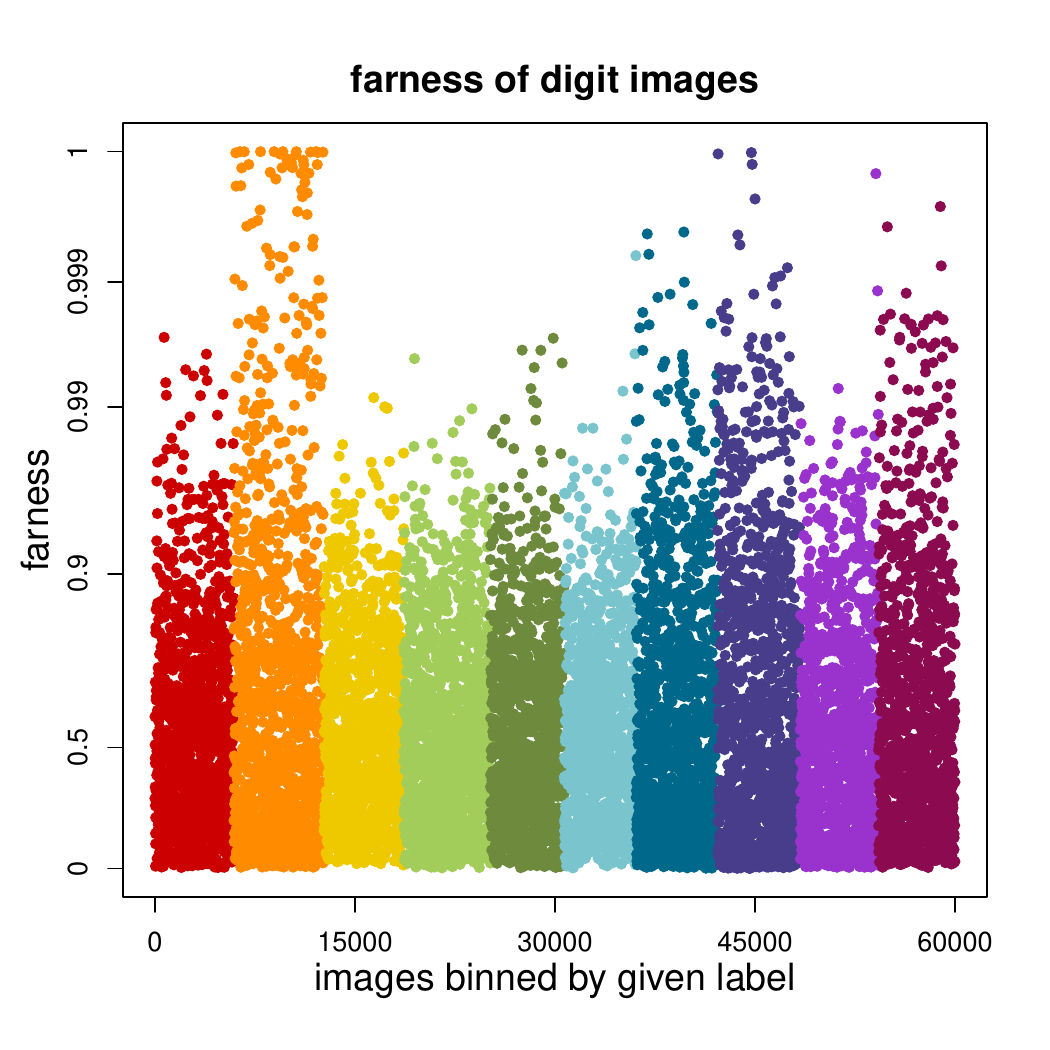}
\vskip-0.2cm
\caption{MNIST data: $\farness$ of each image
         from its given class, binned by class.}
\label{fig:farnessMNIST}
\end{figure}

\newpage
Figure \ref{fig:farnessMNIST} shows the $\farness$ of
each image from its own class, binned by
class (digit). 
Second from left is digit 1, whose $\farness$ has the 
longest tail. We saw in the main text that this class 
indeed has an interesting class map.

\FloatBarrier

\begin{figure}[!ht]
\vspace{0.3cm}
\centering
\includegraphics[width = 0.65\textwidth]
  {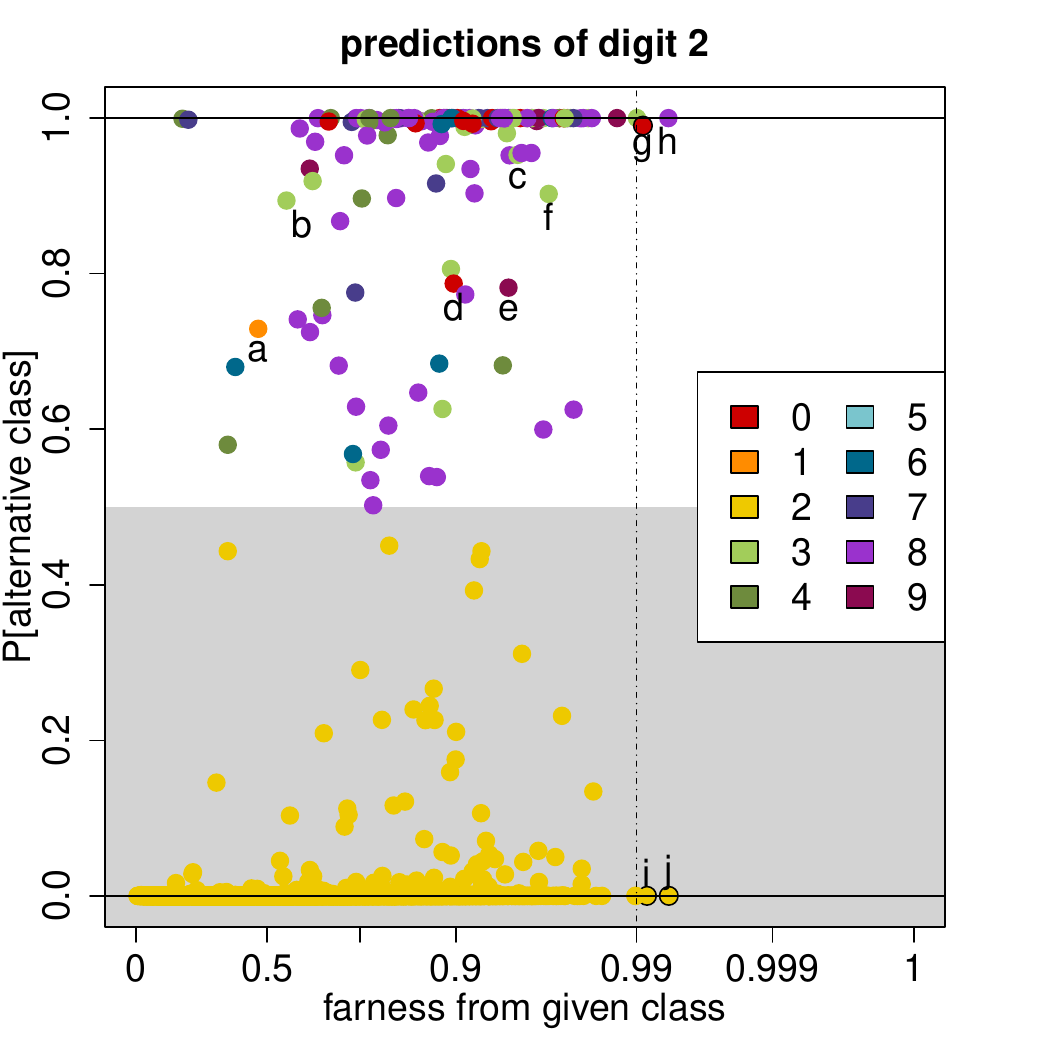}\\
\vspace{0.3cm}
\includegraphics[width = 0.6\textwidth]
  {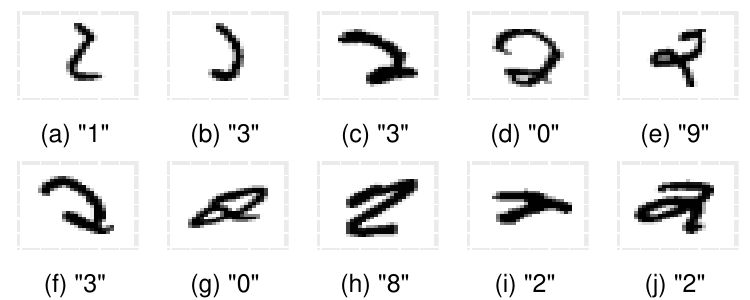}
\caption{Class map of digit 2, with the images 
         corresponding to the marked points.}
\label{fig:MNIST_classmap_digit2}
\end{figure}

Figure~\ref{fig:MNIST_classmap_digit2} shows the 
class map of digit 2, in which most points have 
a favorably low $\PAC$. 
Out of the 5958 images of this digit, 151 were 
misclassified. 
We see that most points have an unexceptional 
$\farness$, indicating that they are relatively
close to class 2, despite some of them being 
misclassified. A few points have been marked for
further inspection. 
Digits \texttt{a} to \texttt{e} are misclassified 
with medium to high conviction.
We see that the corresponding images are very poorly 
written 2's.
Images \texttt{b} and \texttt{c} could be seen as 
the top part of the digit 3, whereas \texttt{d}
is almost a closed circle, explaining its 
misclassification as a 0.  
Digit \texttt{f} also looks like the top part of the 
digit 3, \texttt{g} is basically a zero with an extra 
pen stroke, and \texttt{h} resembles part of an 8.
Finally, points \texttt{i} and \texttt{j} are not
close to any class. Especially image \texttt{i} does
not look like a digit.

\clearpage

\begin{figure}[!ht]
\vspace{0.3cm}
\centering
\includegraphics[width = 0.65\textwidth]
  {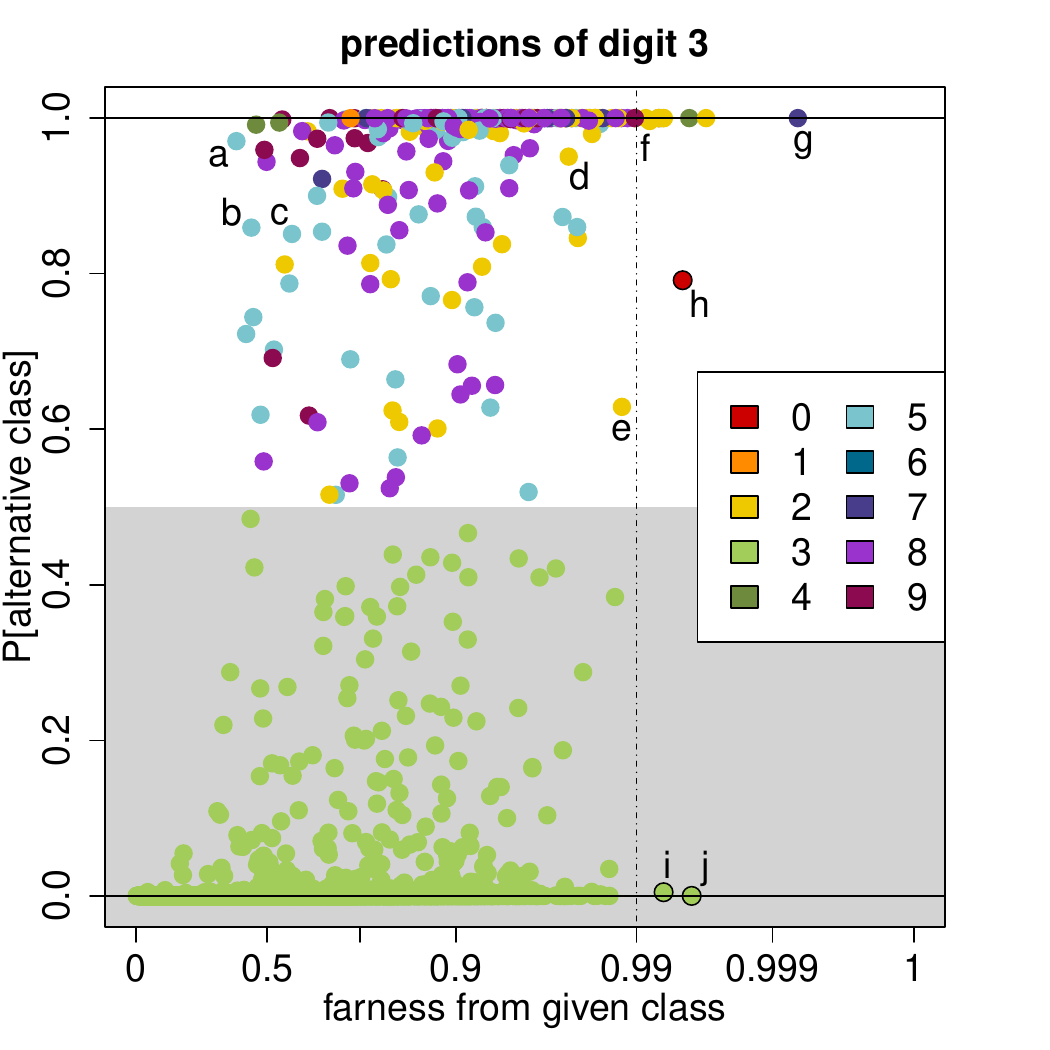}\\
\vspace{0.3cm}
\includegraphics[width = 0.6\textwidth]
  {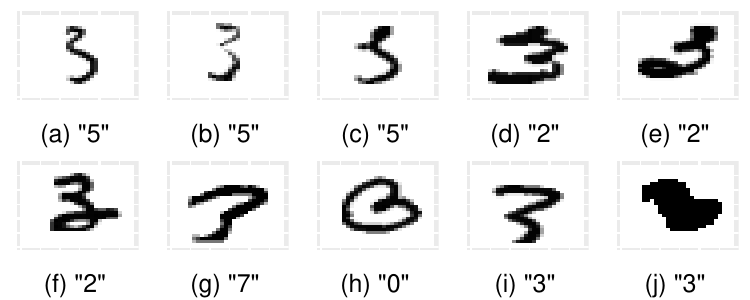}
\caption{Class map of the digit 3, with the images 
  corresponding to the marked points.}
\label{fig:MNIST_classmap_digit3}
\end{figure}

The class map of digit 3 in 
Figure~\ref{fig:MNIST_classmap_digit3} is also 
interesting. Quite a few digits are misclassified
as a 5 instead of a 3, such as \texttt{a}, 
\texttt{b} and \texttt{c}. The images under the 
class map explain why: when the top part of the 
digit 3 is written much smaller than the lower 
part, it indeed bears similarity to a 5. 
Points \texttt{d}, \texttt{e} and \texttt{f} are 
predicted in class 2, as confirmed by their
images. Especially 
images \texttt{e} and \texttt{f}, written with an 
extra curl at the bottom, look as though they are 
a combination of a 2 and a 3. 
Points \texttt{g} to \texttt{j} are outliers,
meaning that they have a high $\farness$ to all
classes. Their images are barely recognizable.
Image \texttt{g} looks like a 7, and \texttt{h} 
is heart-shaped.
Image \texttt{i} is a 3 with a missing bottom,
and image \texttt{j} looks like a stain 
rather than a digit.
The class maps of the remaining digits below
are interpreted analogously.

\begin{figure}[!ht]
\vspace{2.5cm}
\centering
\includegraphics[width = 0.65\textwidth]
  {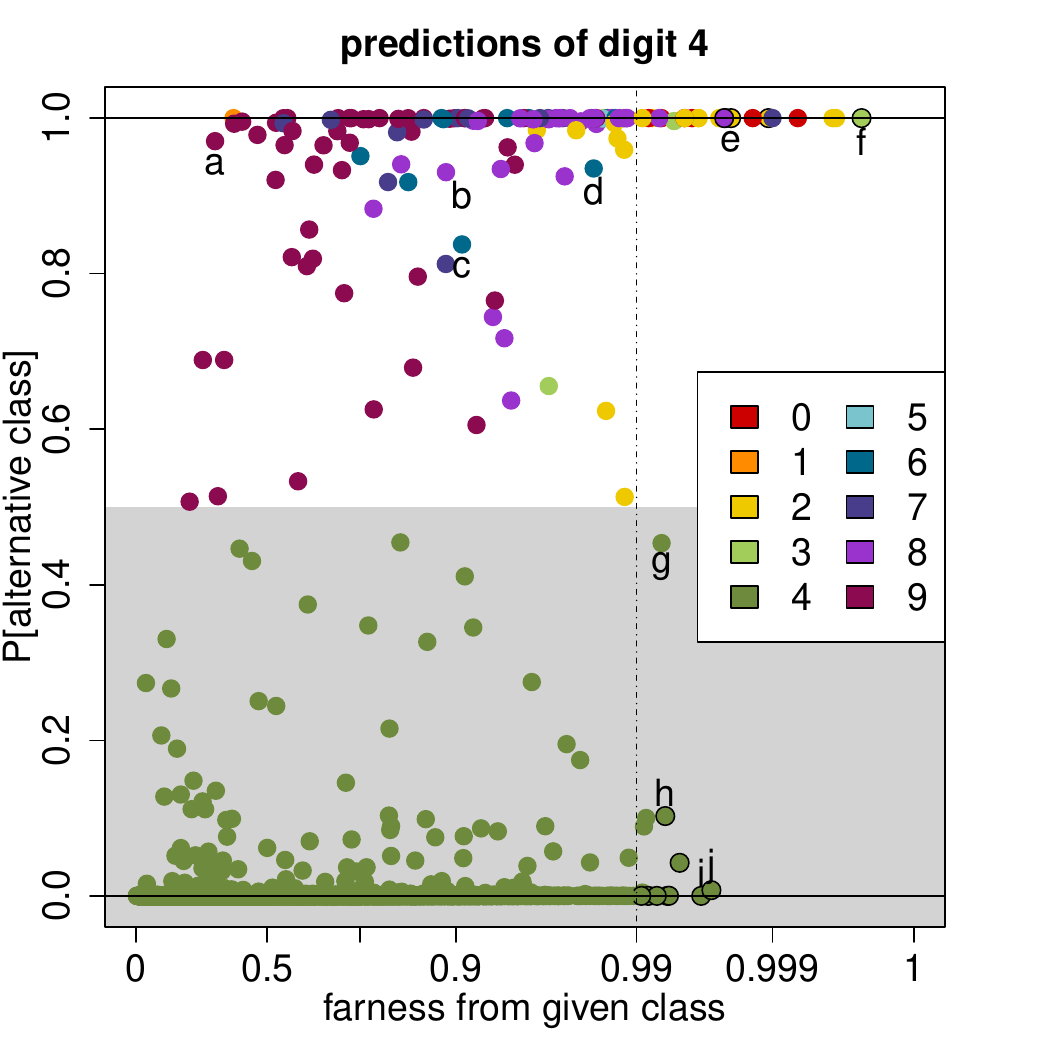}\\
\vspace{0.3cm}
\includegraphics[width = 0.6\textwidth]
  {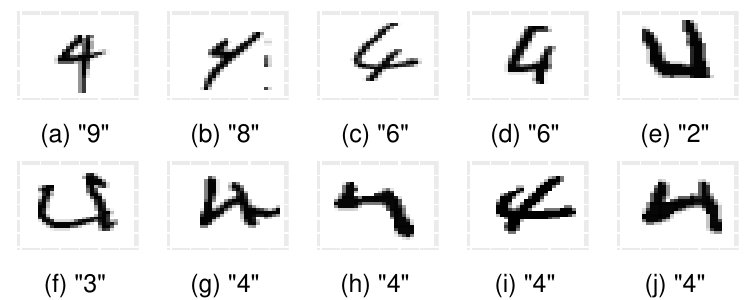}
\caption{Class map of digit 4, with the images 
  corresponding to the marked points.}
\label{fig:MNIST_classmap_digit4}
\end{figure}

\begin{figure}[!ht]
\centering
\includegraphics[width = 0.65\textwidth]
  {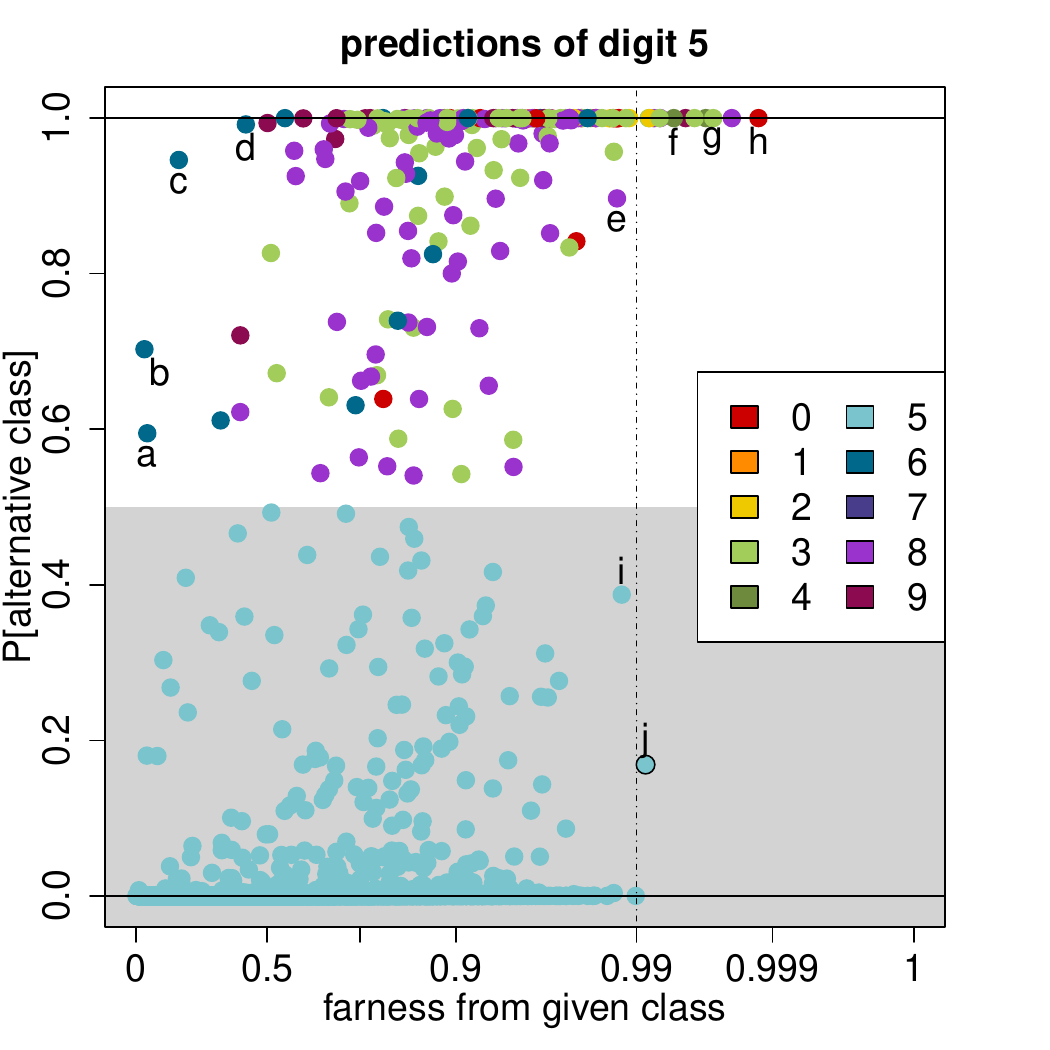}\\
\vspace{0.3cm}
\includegraphics[width = 0.6\textwidth]
  {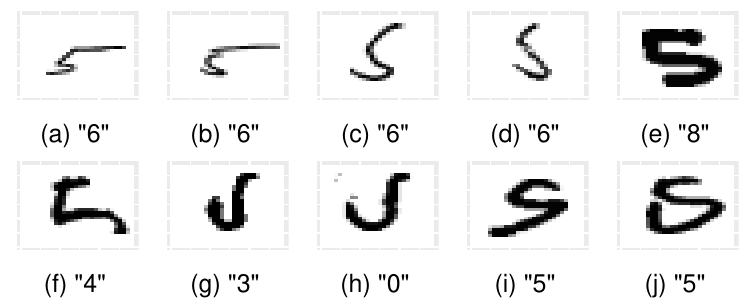}\\
\vspace{0.3cm}
\caption{Class map of digit 5, with the images 
  corresponding to the marked points.}
\label{fig:MNIST_classmap_digit5}
\end{figure}

\begin{figure}[!ht]
\centering
\includegraphics[width = 0.65\textwidth]
  {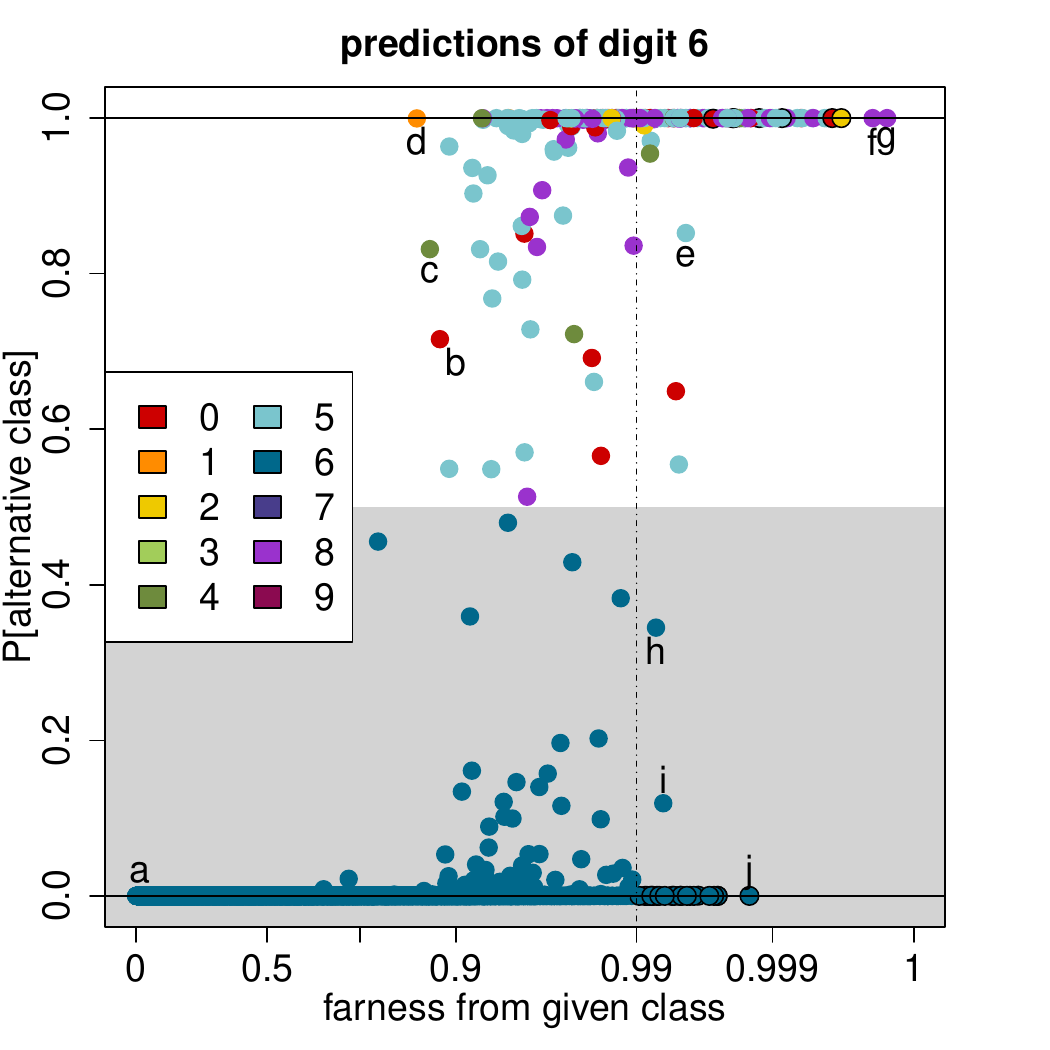}\\
\vspace{0.3cm}
\includegraphics[width = 0.6\textwidth]
  {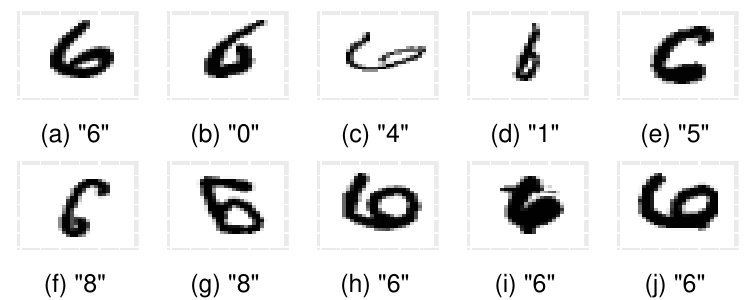}\\
\vspace{0.3cm}	
\caption{Class map of digit 6, with the images 
  corresponding to the marked points.}
\label{fig:MNIST_classmap_digit6}
\end{figure}

\begin{figure}[!ht]
\centering
\includegraphics[width = 0.65\textwidth]
  {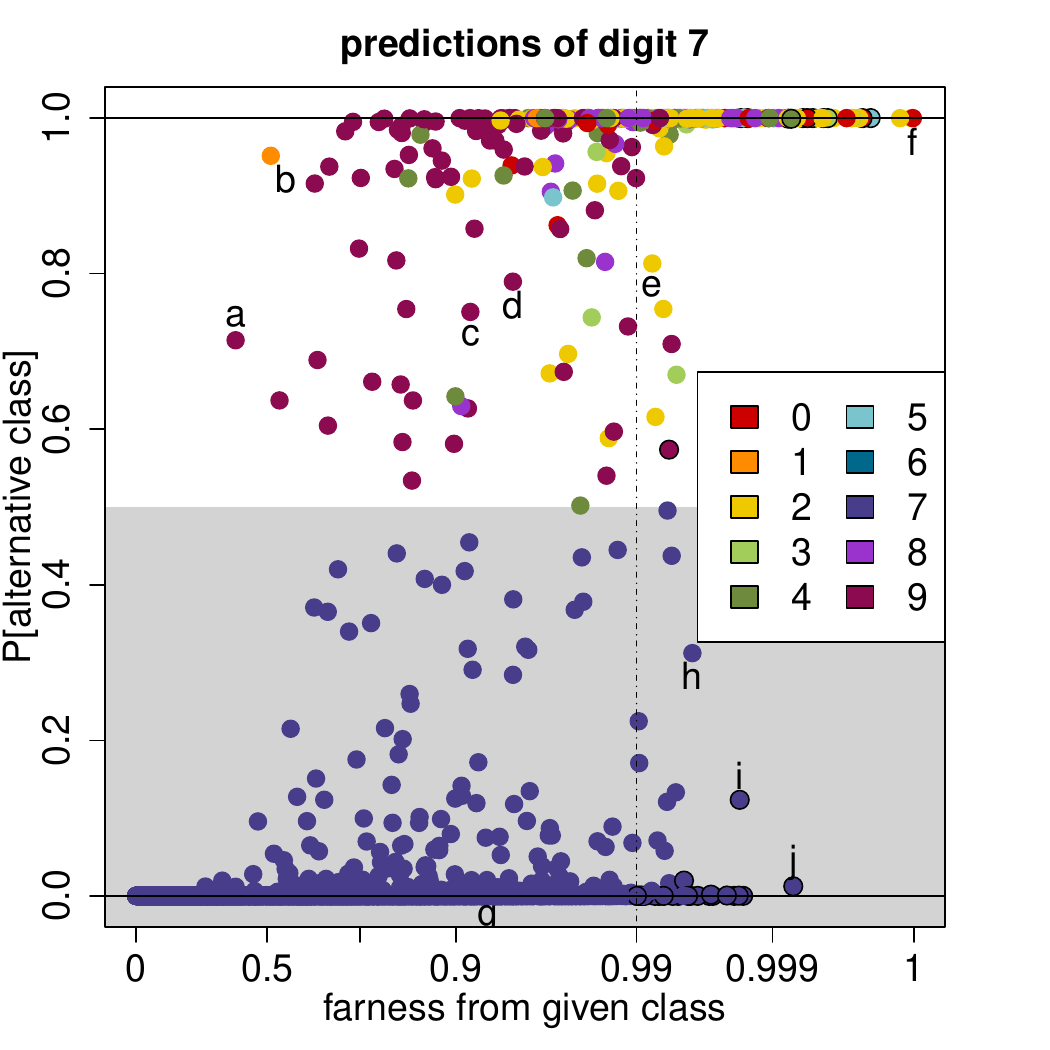}\\
\vspace{0.3cm}
\includegraphics[width = 0.6\textwidth]
  {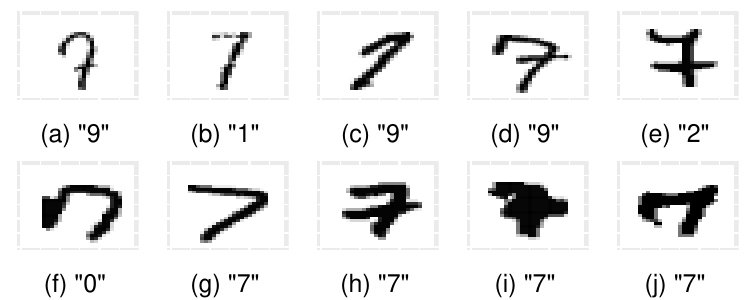}
\caption{Class map of digit 7, with the images 
  corresponding to the marked points.}
\label{fig:MNIST_classmap_digit7}
\end{figure}

\begin{figure}[!ht]
\centering
\includegraphics[width = 0.65\textwidth]
  {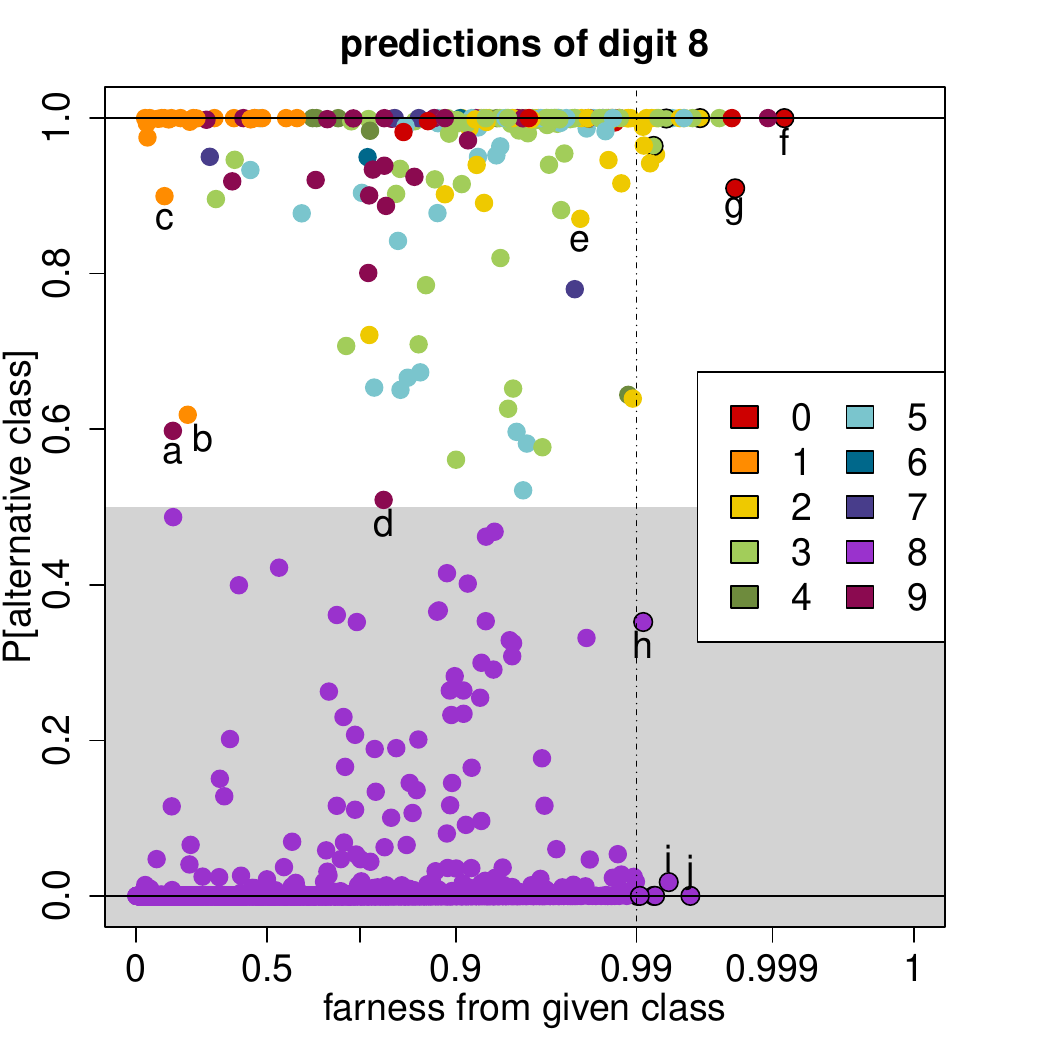}\\
\vspace{0.3cm}
\includegraphics[width = 0.6\textwidth]
  {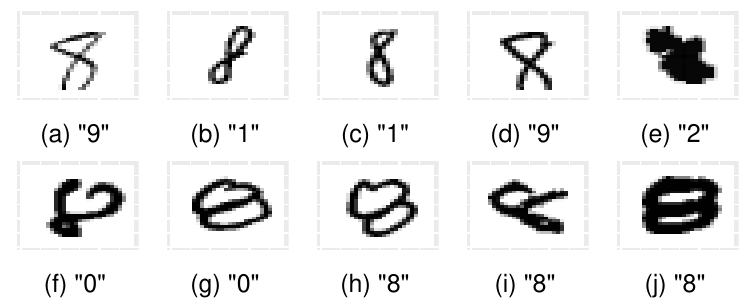}\\
\vspace{0.3cm}	
\caption{Class map of digit 8.}
\label{fig:MNIST_classmap_digit8}
\end{figure}

\begin{figure}[!ht]
\centering
\includegraphics[width = 0.65\textwidth]
  {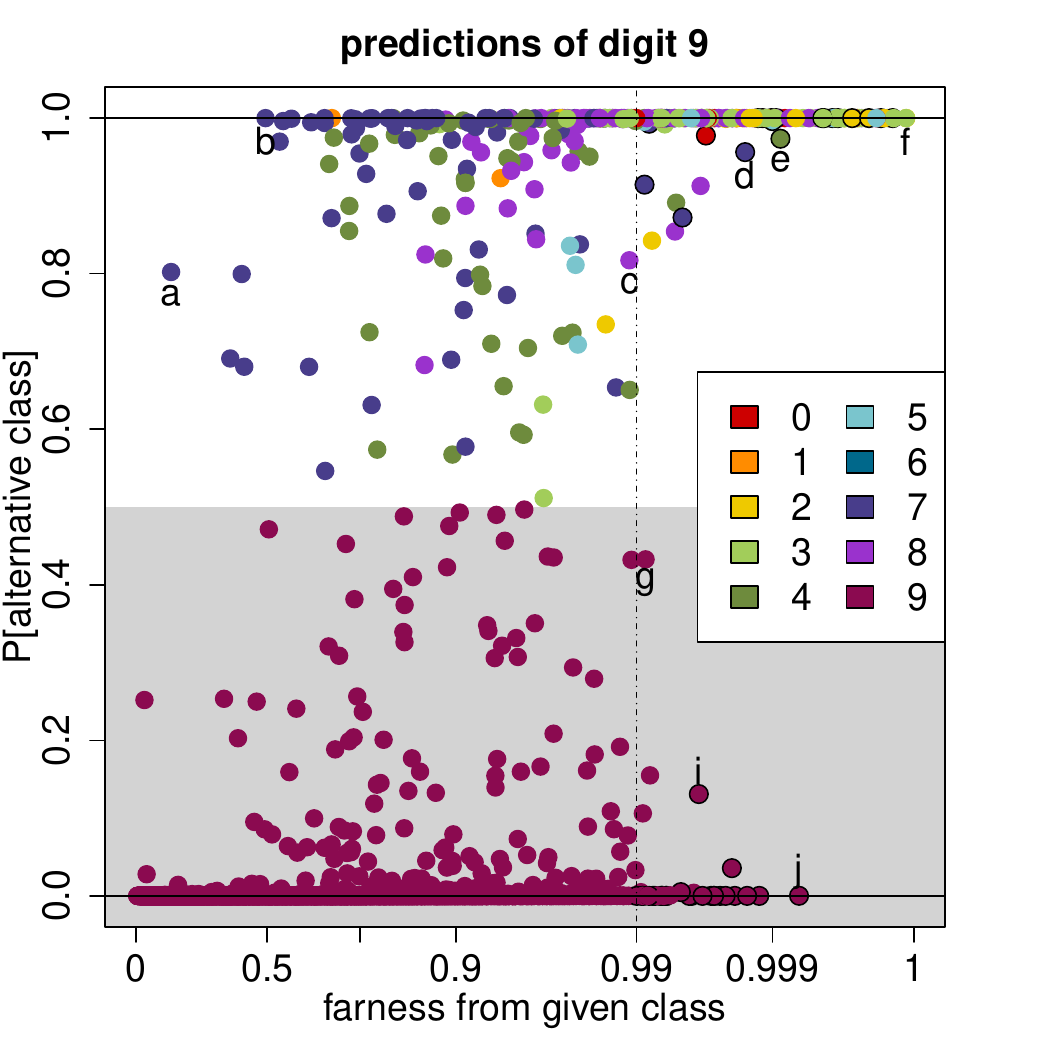}\\
\vspace{0.3cm}
\includegraphics[width = 0.6\textwidth]
  {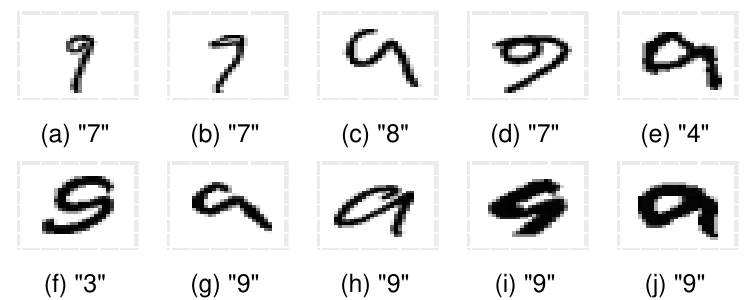}\\
\vspace{0.3cm}	
\caption{Class map of digit 9.}
\label{fig:MNIST_classmap_digit9}
\end{figure}

\clearpage
\subsection{MNIST test data}
\label{suppmat:MNIST_test}

When classifying the MNIST test data using the
trained QDA model, the misclassification rate
is about the same as on the training data.
Therefore the classification is stable, there is 
no indication that the model was overfitted on
the training data.

\begin{figure}[!ht]
\center
\vskip2cm
\includegraphics[width = 0.95\textwidth]
  {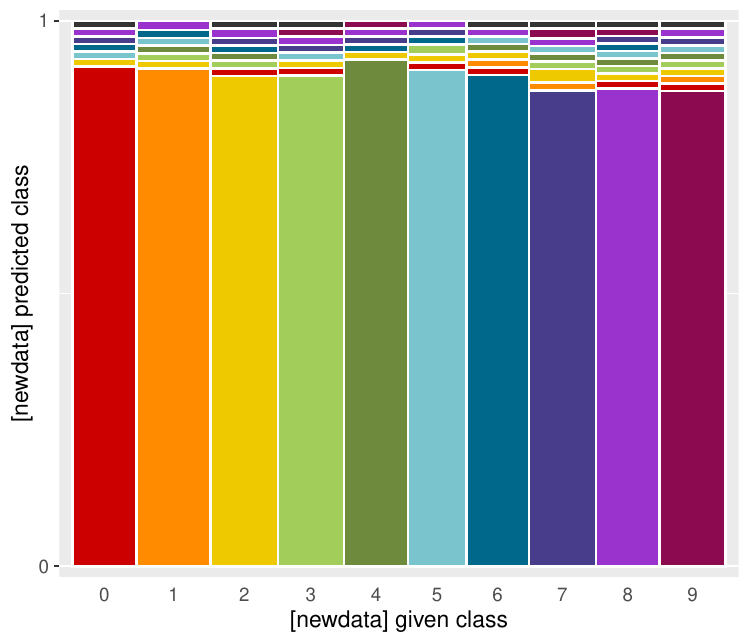}
\caption{Stacked mosaic plot of the MNIST test data.}
\label{fig:stackplot_MNIST_test}
\end{figure}

The stacked mosaic plot of the test data in 
Figure~\ref{fig:stackplot_MNIST_test} is extremely
similar to that of the training data in
Figure~\ref{fig:stackplot_MNIST}.

\begin{figure}[!ht]
\vspace{0.3cm}
\centering
\includegraphics[width = 0.65\textwidth]
  {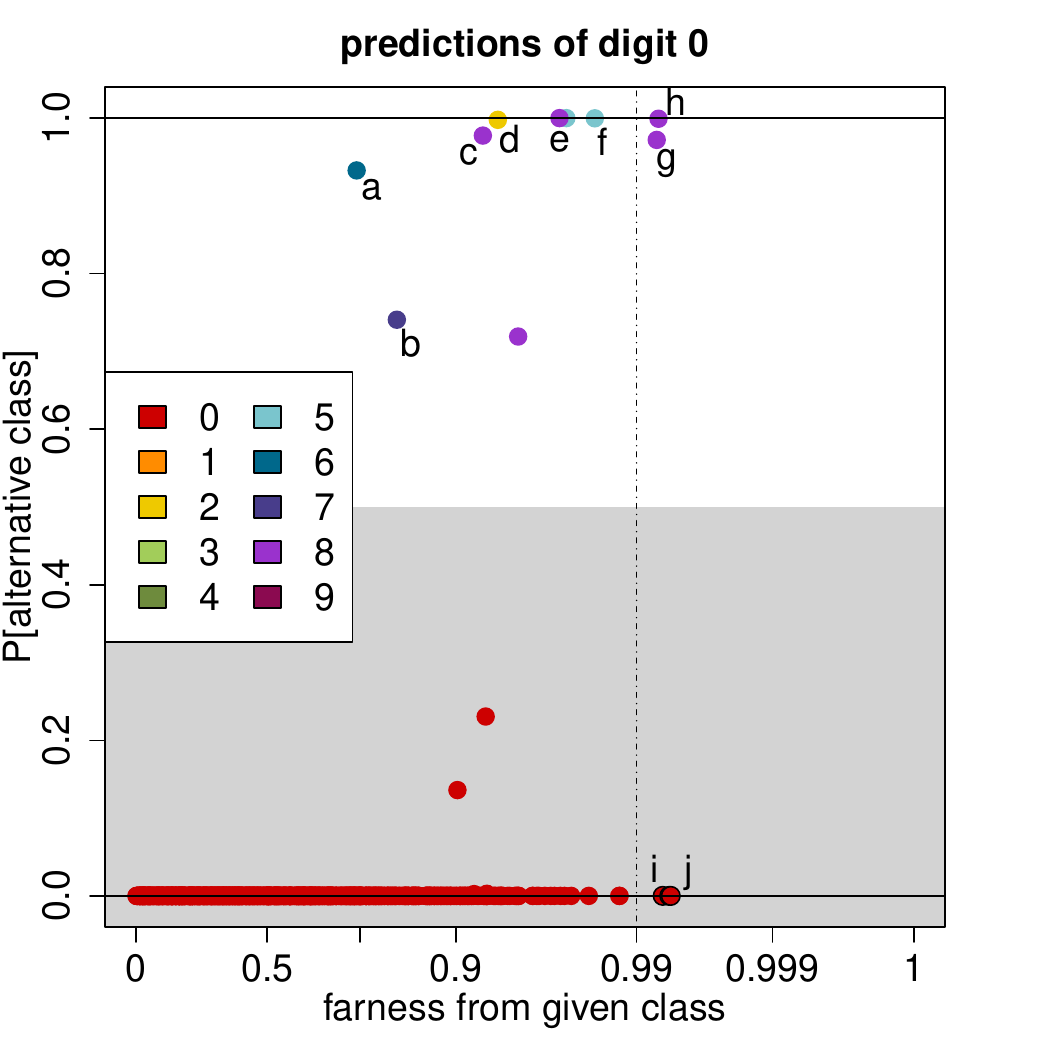}\\
\vspace{0.3cm}
\includegraphics[width = 0.6\textwidth]
  {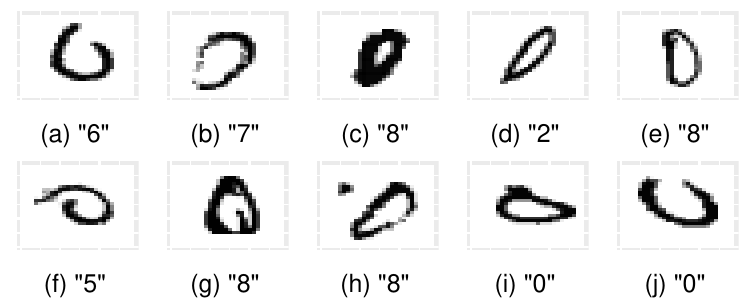}
\caption{Class map of the digit 0, with the images 
  corresponding to the marked points.}
\label{fig:MNISTtest_classmap_digit0}
\end{figure}

\newpage
The class maps of the test data also look
similar to those of the training data.
When comparing the class map of the digit 0 in 
the test data in 
Figure~\ref{fig:MNISTtest_classmap_digit0}
to that of the same digit in the training data in
Figure~\ref{fig:MNIST_classmap_digit0},
we see the main characteristics are the same.
The large majority of points again have 
$\PAC \approx 0$, meaning they lie well within 
the class.
There would seem to be fewer points with
$\PAC > 0$, but that is due to the size of the 
test set, which is 6 times smaller than the 
training set.
As before, a few points are assigned to classes 
0, 2, 6 and some others.

\begin{figure}[!ht]
\vspace{0.3cm}
\centering
\includegraphics[width = 0.65\textwidth]
  {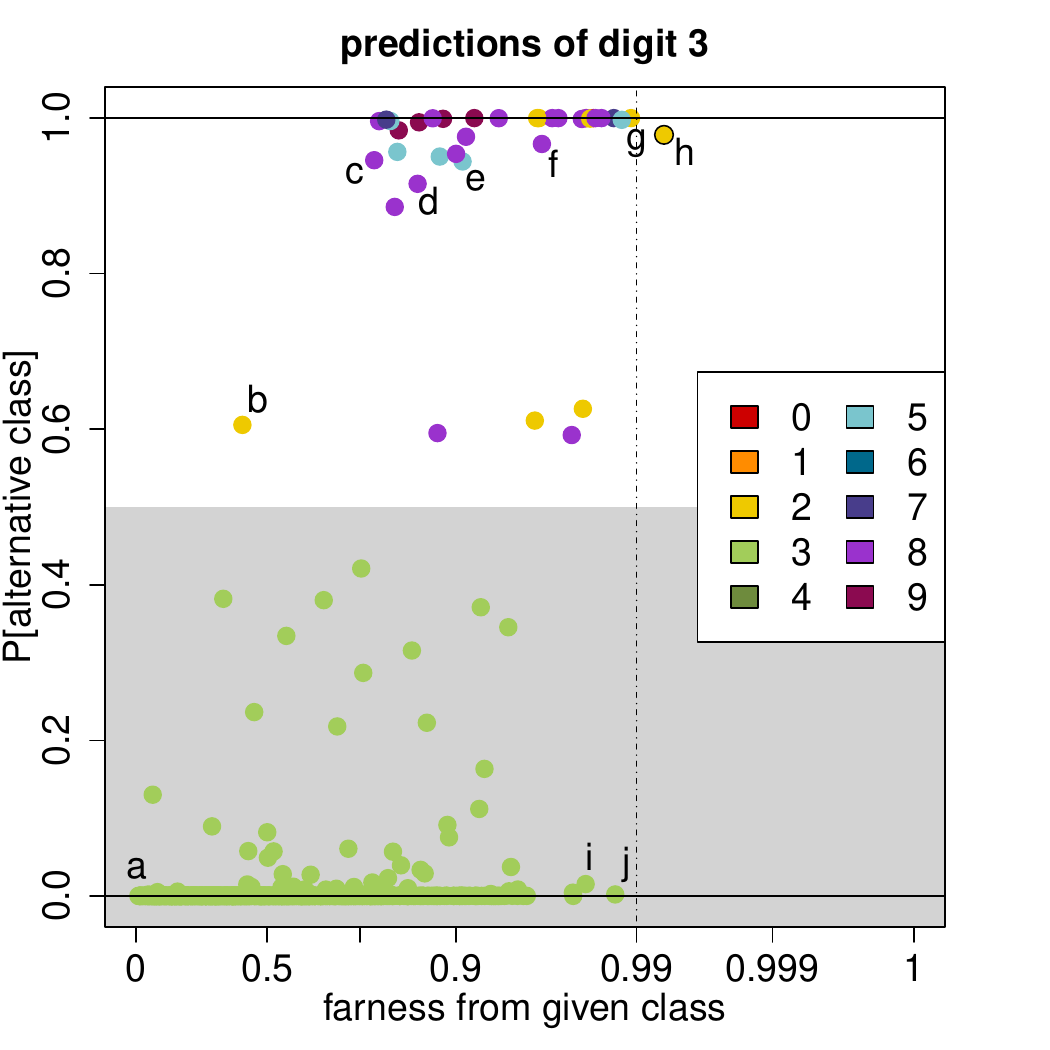}\\
\vspace{0.3cm}
\includegraphics[width = 0.6\textwidth]
  {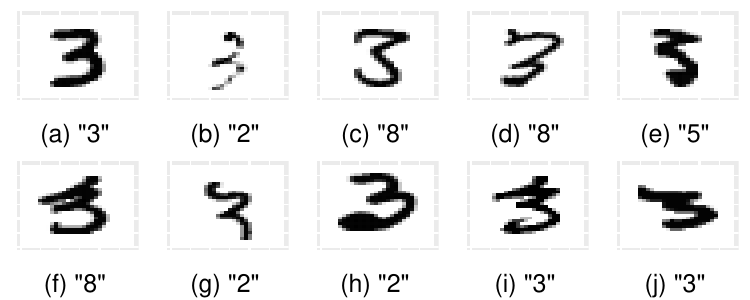}
\caption{Class map of the digit 3, with the images 
  corresponding to the marked points.}
\label{fig:MNISTtest_classmap_digit3}
\end{figure}

\clearpage
Also the class map of digit 3 in 
Figure~\ref{fig:MNISTtest_classmap_digit3}
resembles the one in the training data
(Figure~\ref{fig:MNIST_classmap_digit3}).
Point \texttt{a} has low $\PAC$ and $\farness$,
and its image looks rather perfect.
But overall the $\PAC$ values remain more 
dispersed than with digit 0, with more 
misclassifications.
The images corresponding to the points marked
\texttt{b} to \texttt{h} indicate why they were 
misclassified.
Points \texttt{i} and \texttt{j} are classified
correctly with $\PAC \approx 0$, but look quite
ugly which explains their high $\farness$.

\clearpage
\subsection{Variables in the spam data}
\label{suppmat:spam}

\begin{table}[!ht]
\caption{The variables in the spam dataset}
\begin{tabular}{|p{20mm}|p{85mm}|p{42mm}|}
\hline
Variable number(s) & Variable name(s) & Interpretation\\
\hline
1-48 & \texttt{make}, \texttt{address}, \texttt{all}, 
\texttt{num3d}, \texttt{our}, \texttt{over}, 
\texttt{remove}, \texttt{internet}, \texttt{order}, 
\texttt{mail}, \texttt{receive}, \texttt{will}, 
\texttt{people}, \texttt{report}, \texttt{addresses}, 
\texttt{free}, \texttt{business}, \texttt{email}, 
\texttt{you}, \texttt{credit}, \texttt{your}, 
\texttt{font}, \texttt{000}, \texttt{money}, 
\texttt{hp}, \texttt{hpl}, \texttt{george}, 
\texttt{650}, \texttt{lab}, \texttt{labs}, 
\texttt{telnet}, \texttt{857}, \texttt{data}, 
\texttt{415}, \texttt{85}, \texttt{technology}, 
\texttt{1999}, \texttt{parts}, \texttt{pm}, 
\texttt{direct}, \texttt{cs}, \texttt{meeting}, 
\texttt{original}, \texttt{project}, \texttt{re}, 
\texttt{edu}, \texttt{table}, \texttt{conference} &
percentage of words equal to the given word\\
\hline
49-54 & \texttt{charSemicolon},\linebreak 
\texttt{charRoundbracket},\linebreak 
\texttt{charSquarebracket},\linebreak 
\texttt{charExclamation},\linebreak 
\texttt{charDollar},\linebreak 
\texttt{charHash} & 
fraction of
characters equal to 
 ; ( [  ! \$ \#\\
\hline
55 & \texttt{capitalAve}& average run length 
of\linebreak capital letters\\
\hline
56 & \texttt{capitalLong}& longest run length of 
capital letters\\
\hline	
57 & \texttt{capitalTotal}& total run length of 
capital letters\\
\hline
\end{tabular}
\label{tab:spam_vars}
\end{table}

\newpage
\subsection{More on the book review data}
\label{suppmat:bookreviews}

\begin{figure}[!ht]
\vspace{0.5cm}
\centering
\includegraphics[width = 0.75\textwidth]
  {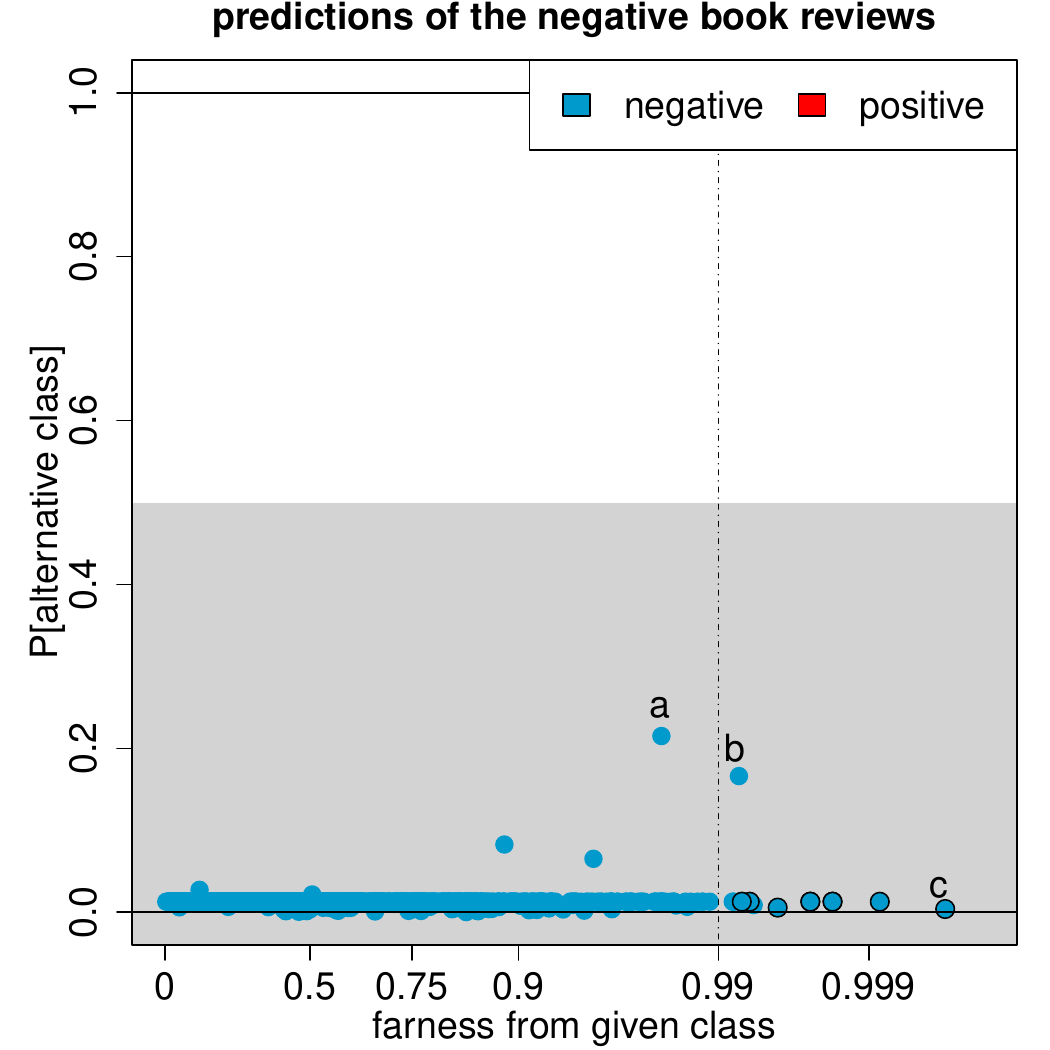}
\caption{Class map of the negative book reviews in 
         the training data.}
\label{fig:bookreview_train_negative}
\end{figure}

Figure \ref{fig:bookreview_train_negative} shows
the class map of the negative reviews in the 
training data.
Most of the $\farness$ values are 
uneventful, and all values of $\PAC$ are small.
Point \texttt{a} stands out a bit because
it has the highest $\PAC$.
Indeed, review \texttt{a} is very positive,
as seen in 
Table~\ref{tab:excerpts_train_negative}.
This review is mislabeled, perhaps the reviewer 
made a mistake when filling in the stars.
Nevertheless, because the classifier was trained
on these data, review \texttt{a} is predicted as 
negative.
Review \texttt{b} has a high $\farness$.
It is correctly predicted as negative. 
The high $\farness$ can be explained by the sheer 
length of this book review. It is over 20,000 
characters long, whereas the next longest review 
in the data is under 10,000 characters and the 
average length of negative reviews is 940 
characters. 
Finally, point \texttt{c} corresponds to
a very short review of only 88 characters,
which explains why it lies far from both groups,
as indicated by its black border.

\begin{table}[!ht]
\centering
\caption{Excerpts of the negative reviews 
         \texttt{a}, \texttt{b} and \texttt{c}.}
\begin{tabular}{|p{13mm}|p{130mm}|}
\hline
marked & excerpts from the book reviews\\
\hline
\texttt{a} &
    ``the left behind series is the best reading 
    i have ever read.'' \newline 
		``when i read the very first book i was 
		hooked''\newline 
		``thank you tim and jerry for such great books''\\
\texttt{b} & 
    ``there is not quite the neglect that nash 
    claimed in these fields'' \newline 
		``nash is not the lone voice for these 
		'forgotten' as he claimed''\\
\texttt{c} & 
		``very disappointed in the contents of this book.
		i expected more\newline information and patterns''\\
\hline
\end{tabular}
\vspace{0.3cm}
\label{tab:excerpts_train_negative}
\end{table}

\begin{figure}[!ht]
\vspace{0.3cm}
\centering
\includegraphics[width = 0.75\textwidth]
  {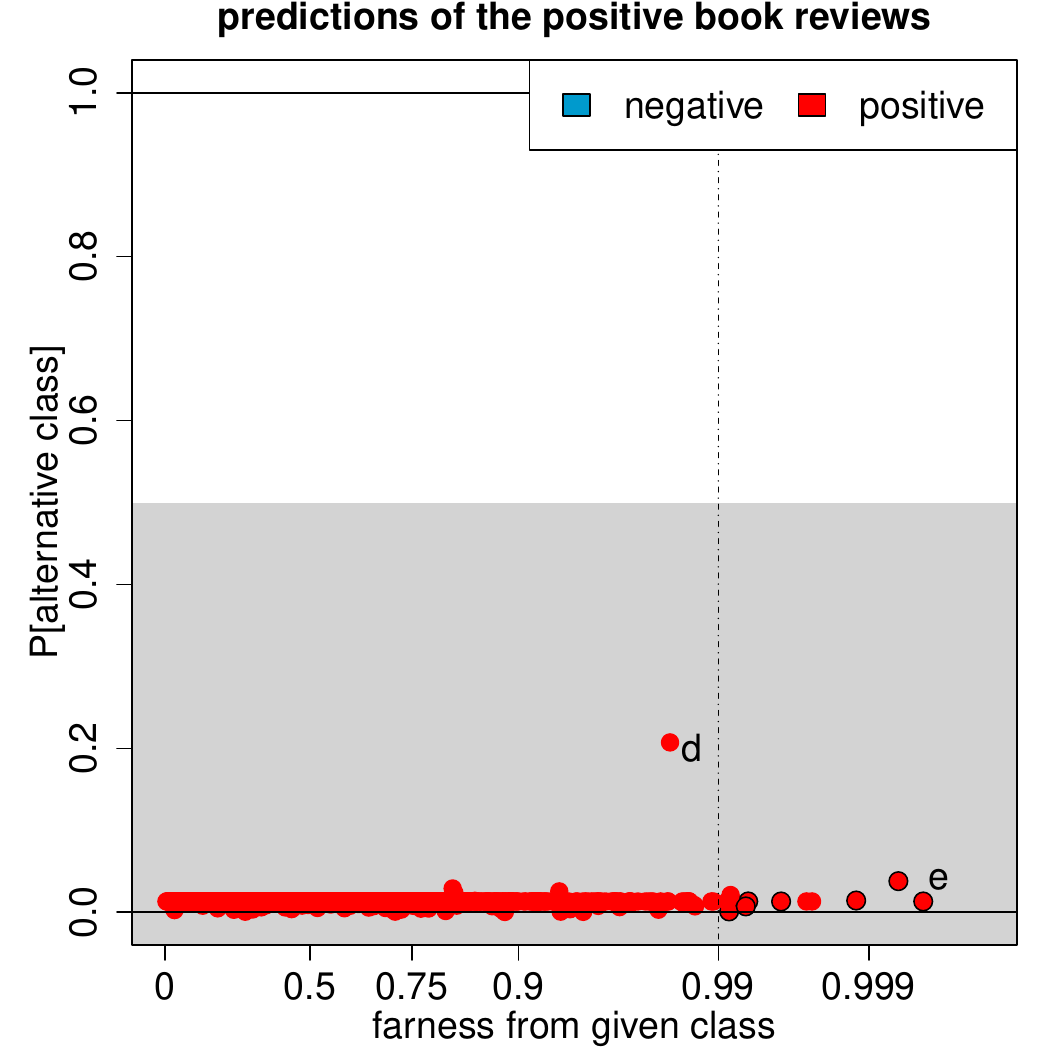}
\caption{Class map of the positive book reviews 
         in the training data.}
\label{fig:bookreview_train_positive}
\end{figure}

Figure \ref{fig:bookreview_train_positive} shows 
the class map of the positive book reviews in the 
training data. We discuss two of the more extreme
points. Review \texttt{d} has the highest
$\PAC$, indicating that the classifier is less
convinced that it is positive. 
It is in fact positive, but it contains many 
negative words due to its rant against other
reviews that were negative about the book.
Review \texttt{e} has a high farness with respect to
both the positive and negative reviews. This 
positive review is very short as it contains only
102 characters, whereas the positive reviews
have an average length of 873 characters.

\begin{table}[!ht]
\centering
\vspace{0.3cm}
\caption{Excerpts of the positive reviews 
         \texttt{d} and \texttt{e}.}
\begin{tabular}{|p{13mm}|p{130mm}|}
\hline
marked & excerpts from the book reviews\\
\hline
\texttt{d} &
    ``i cannot believe the only 2 bad reviews were
     given by people who didn't know the book was 
		 written in spanish'' \newline 
		``if you want to blame somebody because you cannot 
		read the book, blame the editors for not publishing 
		an english version'' \newline 
		``you were careless/stup.. enough to buy a book 
		that you could not read''\\
\texttt{e} &
``very interesting look at the the makeup of the new hampshire \newline
primary and interesting facts and history''\\
\hline
\end{tabular}
\label{tab:excerpts_train_positive}
\end{table}

\clearpage
\subsection{Distance measures used in the paper}
\label{suppmat:distances}

The table below shows the three $D(i,g)$ measures
used in the paper, with the data types and
classifiers for which they can be used.\\

\begin{center}
\includegraphics[height = 19cm] 
  {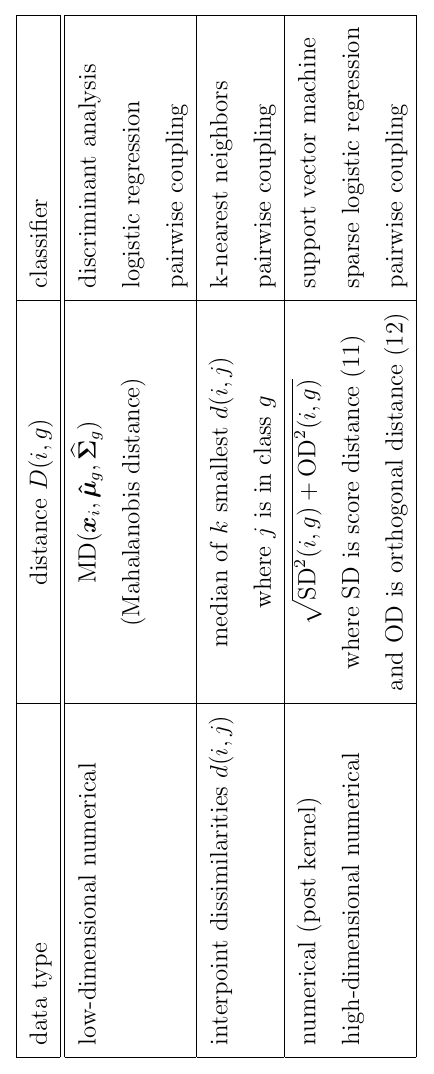}
\end{center}

\clearpage
\subsection{More on the sweets data}
\label{suppmat:sweets}

The table below 
provides the names (in German and French) of the 
products marked in Figure \ref{fig:sweets_4plots} 
in the paper.\\

\begin{table}[ht]
\centering
\small
\begin{tabular}{ll}
  \hline
marked & Name of the product \\ 
  \hline
a & Hot Brownie (Burger King), 1 Stuck = 80g \\ 
  b & Actilife Ballastino Biscuit (Migros) \\ 
  c & Ovomaltine Crunchy Biscuit (Wander) \\ 
  d & Frisco Extreme Mini Mini Chocolat Cornet (Nestle) \\ 
  e & X-Cream Sundae Strawberry (Burger King), 1 Portion = 170g \\ 
  f & Weight Watchers Ice Mokka Becher (Coop) \\ 
  g & Weight Watchers Ice Schokolade Becher (Coop) \\ 
  h & Leger Kakao Glace (Migros) \\ 
  i & Leger Mini Vanille Glace-Stangel (Migros) \\ 
  j & Weight Watchers Ice Camarello mit Caramelaroma Becher (Coop) \\ 
  k & Weight Watchers Ice Exotic Becher (Coop) \\ 
  l & Weight Watchers Lutscher Mini Exotic Ice (Coop) \\ 
  m & Erdbeer-Torte, zubereitet (Dr. Oetker) \\ 
  n & Qualite\&Prix Backmischung Rueblitorte (Coop) \\ 
  o & Weight Watchers Linzertortli (Coop) \\ 
  p & Leisi Cake Chocolat, Flussigteig (Nestle) \\ 
  q & Duo Mousse Chocolat (Migros) \\ 
  r & Dessert Tradition Fondant au Chocolat (Migros) \\ 
  s & Milchreis klassisch, Fertigmischung (Migros) \\ 
  t & Griessbrei Fertigmischung (Migros) \\ 
  u & Varieta Basis-Creme-Pulver (Migros) \\ 
  v & Pudding Creme Vanille ohne Zucker, Fertigmischung (Migros)\\ 
  w & Dawa Flan Caramel, Pulver (Wander) \\ 
  x & Mousse au Chocolat zartbitter (Dr. Oetker), zubereitet \\ 
   \hline
\end{tabular} 
\label{tab:sweets}
\end{table}

\end{document}